\documentclass{bmvc2k}
\usepackage{epsfig}
\usepackage{amsmath}
\usepackage{amssymb}
\usepackage{graphicx}
\usepackage{wrapfig}
\usepackage{caption}
\usepackage{adjustbox}
\usepackage{booktabs}
\usepackage{enumitem}
\usepackage{multirow}
\usepackage{makecell}
\usepackage{url}
\usepackage[dvipsnames]{xcolor}
\usepackage{subcaption}

\title{AI-Generated Image Detection: An Empirical Study and Future Research Directions}
\vspace{-20pt}
\addauthor{Nusrat Tasnim}{tasnim.nishu70@kau.kr}{1}
\addauthor{Kutub Uddin}{kutub@umich.edu}{2}
\addauthor{Khalid Mahmood Malik}{drmalik@umich.edu}{2}

\addinstitution{
 School of Electronics and Information Engineering\\
 Korea Aerospace University\\
 Goyang, South Korea
}
\addinstitution{
 College of Innovation \& Technology\\
 University of Michigan-Flint,\\
 Michigan, USA
}
\vspace{-20pt}

\runninghead{Tasnim et al.}{AI-Generated Image Detection: An Empirical Study}

\begin{document}

\maketitle
\vspace{-20pt}
\begin{abstract}
The threats posed by AI-generated media, particularly deepfakes, are now raising significant challenges for multimedia forensics, misinformation detection, and biometric system resulting in erosion of public trust in the legal system, significant increase in frauds, and social engineering attacks.
Although several forensic methods have been proposed, they suffer from three critical gaps: (i) use of non-standardized benchmarks with GAN- or diffusion-generated images, (ii) inconsistent training protocols (e.g., scratch, frozen, fine-tuning), and (iii) limited evaluation metrics that fail to capture generalization and explainability. These limitations hinder fair comparison, obscure true robustness, and restrict deployment in security-critical applications.
This paper introduces a unified benchmarking framework for systematic evaluation of forensic methods under controlled and reproducible conditions. We benchmark ten SoTA forensic methods (scratch, frozen, and fine-tuned) and seven publicly available datasets (GAN and diffusion) to perform extensive and systematic evaluations. We evaluate performance using multiple metrics, including accuracy, average precision, ROC-AUC, error rate, and class-wise sensitivity. We also further analyze model interpretability using confidence curves and Grad-CAM heatmaps.
Our evaluations demonstrate substantial variability in generalization, with certain methods exhibiting strong in-distribution performance but degraded cross-model transferability. This study aims to guide the research community toward a deeper understanding of the strengths and limitations of current forensic approaches, and to inspire the development of more robust, generalizable, and explainable solutions.
\end{abstract}
\vspace{-20pt}
\section{Introduction}\vspace{-8pt}
\label{sec:intro}
In recent times, the proliferation of AI-generated content, particularly deepfakes~\cite{uddin2024counter}, has overwhelmed social media~\cite{DeepfakesSocialMedia} and news platforms~\cite{DeepfakeNews}. These deepfake contents are often used to mislead audiences by fabricating events or impersonating individuals, thereby undermining public trust. Moreover, deepfakes have emerged as critical threats to society, especially in security-sensitive domains. For instance, they can compromise biometrics used for face recognition and identification~\cite{BiometricThreat, uddin2019anti}, surveillance systems~\cite{Surveillance}, and mislead perception modules in autonomous driving~\cite{AutonomousDriving, tasnim2023dynamic}. Additionally, deepfakes pose significant risks in the Internet of Things (IoT) ecosystem~\cite{IoT, tasnim2020deep} and remote authentication systems~\cite{RemoteAccess, tasnim2022deep}, where identity integrity is crucial. As the quality and realism of AI-generated content continue to improve, the ability to detect deepfake media has become increasingly challenging and making it imperative to develop robust and generalizable techniques.\\
Current statistics highlight the urgent need to verify deepfake content in domains such as social media, journalism, finance, the legal system, and governance. Across industries, deepfake fraud has become alarmingly common. Nearly 92\% of companies have reported financial losses due to deepfake scams~\cite{cfo2024}. On average, businesses lose approximately \$450,000 per incident, with the financial sector bearing even heavier losses, averaging \$600,000 per organization, and in some cases exceeding \$1 million~\cite{globenewswire2024}. In one notable incident, a Hong Kong employee was tricked into transferring \$25 million after fraudsters used a deepfake video call to impersonate the company’s CFO~\cite{eftsure2024}.\\
Globally, deepfake-enabled fraud caused over \$200 million in losses during the first quarter of 2025 alone, indicating a rapidly escalating threat~\cite{esecurityplanet2025}. The cryptocurrency sector saw an even more dramatic impact, with deepfake-related scams increasing by 456\% between May 2024 and April 2025, culminating in more than \$10.7 billion in damages in 2024~\cite{reddit2025}. Market projections suggest that generative AI-related fraud losses could rise to \$40 billion by 2027, up from \$12.3 billion in 2023~\cite{eftsureForecast}. Beyond financial harm, the reputational damage from deepfakes is equally concerning. Victims often suffer from long-term erosion of trust, reputational fallout, and brand damage. In one instance, a fabricated image of an explosion near the Pentagon, generated by AI, caused a temporary dip in the Dow Jones index, highlighting how deepfakes can disrupt public confidence and financial stability~\cite{realitydefender2023}.\\
These statistics emphasize the urgent need for generalized forensic methods to detect deepfake content, thereby protecting individuals, institutions, and the public at large. Several methods~\cite{wang2020cnn, frank2020leveraging, uddin2023robust, tan2024rethinking, ojha2023towards} have been developed using deep learning~\cite{wang2020cnn} and hybrid~\cite{ojha2023towards} approaches to ensure media integrity. Some approaches~\cite{wang2020cnn, tan2024rethinking} incorporate preprocessing and data augmentation techniques to enhance generalization, while others~\cite{ojha2023towards} leverage SoTA foundation models for robust feature extraction to ensure better generalizability. \\
The major drawbacks of these methods~\cite{wang2020cnn, frank2020leveraging, tan2024rethinking, ojha2023towards} are specific to datasets~\cite{wang2020cnn}, generative models~\cite{frank2020leveraging, uddin2025guard}, or training strategies~\cite{tan2024rethinking, uddin2025shield}. Although many benchmark datasets~\cite{tan2024rethinking} are now publicly available for evaluating a model’s effectiveness and generalizability, most existing approaches~\cite{wang2020cnn, frank2020leveraging, tan2024rethinking, ojha2023towards} consider only one or two datasets for evaluation, leaving many others unexplored. Furthermore, these methods are not assessed within a unified framework, which hinders the reproducibility of results and limits future research. \\
In this article, we conduct an empirical study of generalizable and explainable deepfake detection.  The major contributions are listed as follows:\\ \vspace{-13pt}
\begin{itemize}[noitemsep,topsep=0pt]
\item We propose a unified benchmarking framework that systematically evaluates the generalization capabilities of SoTA forensic methods across benchmark datasets, generative models, and training paradigms.
\item We conduct an extensive empirical study involving ten SoTA detection methods (scratch, frozen, and finetuned) and seven publicly available deepfake datasets (GAN and Diffusion) that offer a comprehensive and reproducible evaluation setup.
\item We incorporate explainability techniques (confidence, ROC curves, and GradCAM) to interpret model predictions and highlight the decisions made by them.
\item We provide critical insights into the strengths and limitations of current forensic methods and identify open challenges that aim to guide the development of more robust, generalizable, and explainable deepfake detection methods.
\end{itemize}
\vspace{-20pt}

\section{Empirical Study Design}
\begin{figure*}[!t]%
\centering
    {\includegraphics[width=1\linewidth]{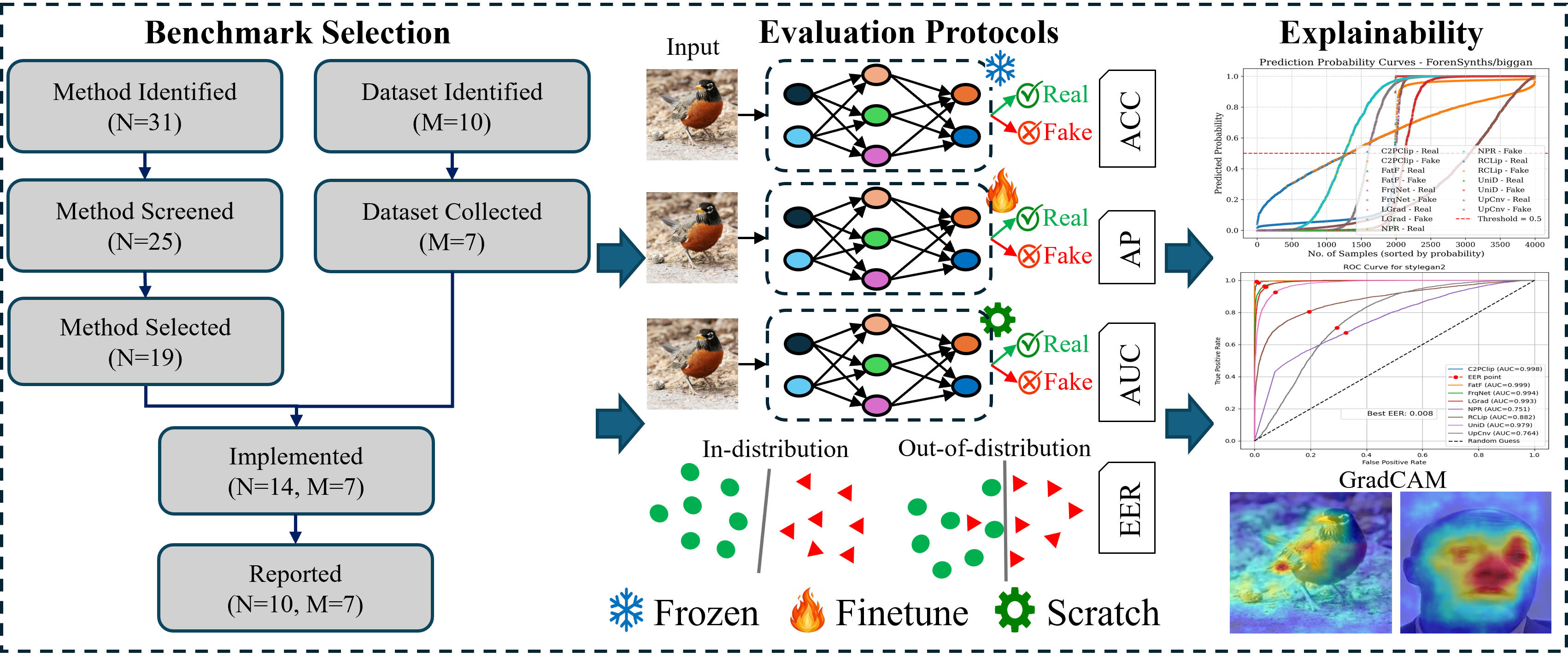}}
    \vspace{0.1pt}
    \caption{Overview of the empirical study: First, we perform benchmark selection, including datasets, forensic, and explainability techniques. Next, we define evaluation protocols covering frozen, fine-tuned, and from-scratch models for AI-generated image detection. Finally, we provide a comprehensive explanation based on confidence, ROC curves, and GradCAM.}
    \label{fig:arch}
    \vspace{-10pt}
\end{figure*}

\vspace{-10pt}
\label{sec:benchmarks}
This section outlines the overall design of the empirical comparative study depicted in Figure~\ref{fig:arch}, covering benchmark selection, evaluation protocols, and explainability techniques. We identified 31 forensic methods and 10 benchmark datasets. Among the 31 forensic methods, we screened 25 and selected 19 based on venue, effectiveness, and novelty. We implemented 14 forensic methods. Similarly, we identified 10 benchmark datasets and collected 7 of them based on accessibility and representation of recent generative models. Owing to the limited generalization ability of the 4 implemented methods and their outdated nature, we ultimately reported generalization and explainability results using 10 forensic methods across 7 datasets, as listed in Table~\ref{tab:dataset_summary} and Table~\ref{tab:methods}. \vspace{-18pt}

\subsection{Datasets}\vspace{-8pt}
This section provides an overview of SoTA benchmark datasets, including their names, release years, object categories, and generative techniques, as summarized in Table~\ref{tab:dataset_summary}.
\vspace{-15pt}

\begin{table*}[!t]
\centering
\small
\renewcommand{\arraystretch}{1.1}
\setlength{\tabcolsep}{5pt}
\caption{Summary of Benchmark Datasets Commonly Used in the Research Community} 
\vspace{4pt}
\begin{adjustbox}{width=1\linewidth}
\begin{tabular}{lcl}
\hline
\textbf{Name} & \textbf{Year} & \textbf{Generative Technique}  \\
\hline
ForenSyn~\cite{wang2020cnn} & 2020 & 
\makecell[l]{ProGAN~\cite{karras2017progressive}, StyleGAN~\cite{karras2019style}, BigGAN~\cite{brock2018large}, CycleGAN~\cite{zhu2017unpaired}, StarGAN~\cite{choi2018stargan}, GauGAN~\cite{park2019semantic}, \\StyleGAN2~\cite{karras2020analyzing}, Deepfakes~\cite{rossler2019faceforensics}} \\
\hline
ForenSynthsCh~\cite{wang2020cnn} & 2022 & 
\makecell[l]{CRN~\cite{chen2017photographic}, IMLE~\cite{li2019diverse}, SAN~\cite{dai2019second}, SIDT~\cite{chen2017photographic}, WFR~\cite{wang2020cnn}} \\
\hline
Diffusion1KStep~\cite{tan2024rethinking} & 2023 & 
Dalle~\cite{ramesh2021zero}, DDPM~\cite{ho2020denoising}, Guided-Diffusion~\cite{dhariwal2021diffusion}, Improved-Diffusion~\cite{nichol2021improved}, Mid-Journey~\cite{tan2024rethinking} \\
\hline
DIRE~\cite{wang2023dire} & 2024 & 
\makecell[l]{ADM~\cite{dhariwal2021diffusion}, DDPM~\cite{ho2020denoising}, IDDPM~\cite{nichol2021improved}, LDM~\cite{rombach2022high}, PNDM~\cite{liu2022pseudo}, SDV1~\cite{rombach2022high}, \\SDV2~\cite{rombach2022high}, VQDiffusion~\cite{gu2022vector}} \\
\hline
GAN~\cite{tan2024rethinking} & 2024 & 
\makecell[l]{AttGAN~\cite{he2019attgan}, BEGAN~\cite{berthelot2017began}, CramerGAN~\cite{bellemarecramer}, InfoMaxGAN~\cite{lee2021infomax}, \\MMDGAN~\cite{li2017mmd},  
RelGAN~\cite{nie2018relgan}, S3GAN~\cite{luvcic2019high}, SNGAN~\cite{miyato2018spectral}, STGAN~\cite{liu2019stgan}} \\
\hline
UClipiffusion~\cite{ojha2023towards} & 2023 & 
\makecell[l]{Dalle~\cite{ramesh2021zero}, Glide (50\_27, 100\_10, 100\_27)~\cite{nichol2021glide}, Guided~\cite{dhariwal2021diffusion}, \\ 
LDM (100, 200, 200\_cfg)~\cite{rombach2022high}} \\
\hline
MNW~\cite{MNW2025} & 2025 & 
\makecell[l]{Adobe, Adversarial, Amazon\_v2, Aura\_flow, Baidu, Bytedance\_v3, Civitai\_v6, Flux, Google, \\Hunyuandit, Hypersd, Ideogram, Kandinsky, Krea\_1, Kuaishou, Luma\_photon, Lumina, \\Meta\_imagine, Midjourney, Nvidia\_sana, Openai, Pixart\_alpha\_xl, Playgroundai, Recraft\_v3, \\Reve\_ai, Stable, Ultrapixel, Wuerstchen} \\
\hline
\end{tabular}
\end{adjustbox}
\label{tab:dataset_summary}
\vspace{-20pt}
\end{table*}

\begin{table*}[ht]
\centering
\small
\renewcommand{\arraystretch}{1.1}
\setlength{\tabcolsep}{5pt}
\caption{Summary of Benchmark Detection Methods Used in the Research Community}
\vspace{4pt}
\begin{adjustbox}{width=1\linewidth}
\begin{tabular}{lcll}
\hline
\textbf{Name} & \textbf{Year} & \textbf{Strength} & \textbf{Limitation} \\
\hline
CNND~\cite{wang2020cnn} & CVPR, 2020 & 
\makecell[l]{Improved generalization by \\ careful data augmentation} & 
\makecell[l]{Did not explore other types of generative \\ models such as diffusion} \\
\hline
LGrad~\cite{tan2023learning}  & CVPR, 2023 &   
\makecell[l]{Extracted gradient features using \\pretrained generative models} & 
\makecell[l]{Limited to StyleGAN and ProGAN models \\ Did not explore any diffusion models} \\
\hline
NPR~\cite{tan2024rethinking}  & CVPR, 2024 & Explored neighboring pixel-relationship & Bias towards the deepfake class\\
\hline
UClip~\cite{ojha2023towards}  & CVPR, 2023 & Does not require retraining of CLIP & Limited generalization to recent generative models\\
\hline
RClip~\cite{cozzolino2024raising} & CVPR, 2024 & Requires less training data & Limited to certain generative models \\
\hline
FatF~\cite{liu2024forgery} & CVPR, 2024 & Captures frequency domain artifacts & 
\makecell[l]{Lacks generalization to recent unseen datasets} \\
\hline
RINE~\cite{koutlis2024leveraging}  & ECCV, 2024 & \makecell[l]{Trainable importance estimator for encoder} & Evaluated on fewer datasets \\
\hline
UpConv~\cite{durall2020watch}  & CVPR, 2020 & Captured spectral features & Comparatively less effective and less generalizable \\
\hline
FreqNet~\cite{tan2024frequency}  & AAAI, 2024 & End-to-end frequency learning model & Does not account for other image properties \\
\hline
C2PClip~\cite{c2pclip2025}  & AAAI, 2025 & \makecell[l]{Caption generation and enhancement \\ Concept injection to finetune CLIP}&  \makecell[l]{Limited analysis of the captions results in incomplete \\information}\\
\hline
\end{tabular}
\end{adjustbox}
\label{tab:methods}
\vspace{-18pt}
\end{table*}

\subsubsection{GAN-Based Datasets}\vspace{-8pt} 
The \textbf{ForenSyn~\cite{wang2020cnn}} dataset was introduced by Wang et al. to improve the generalization capability of generic deepfake detection. It comprises data from eight GAN sources, including three conditional GANs~\cite{choi2018stargan, park2019semantic, zhu2017unpaired}, unconditional GANs~\cite{karras2017progressive, karras2019style}, and a deepfake face~\cite{rossler2019faceforensics} source. Most SoTA methods train their models on the ProGAN~\cite{karras2017progressive} training set. 
\textbf{GAN~\cite{tan2024rethinking}} contains data from 9 GAN sources with varying architectural properties. These data differ from ForenSyn~\cite{wang2020cnn}, which covers a diverse range of wild scenes. In ForenSyn~\cite{wang2020cnn}, each sub-dataset has a random number of real and deepfake samples, while in GAN~\cite{tan2024rethinking}, each sub-dataset comprises 2K real and 2K deepfake images.
\vspace{-18pt}

\subsubsection{Diffusion-based Datasets }\vspace{-8pt}
\textbf{DIRE~\cite{wang2023dire}} consists of 8 diffusion-generated deepfake samples. The real images are randomly collected from ForenSyn~\cite{wang2020cnn} (LSUN~\cite{wang2020cnn} and ImageNet~\cite{russakovsky2015imagenet}) real classes.
\textbf{Diffusion1kStep~\cite{tan2024rethinking}} is a diffusion-family dataset containing data from five diffusion sources. All samples are generated using 1K diffusion steps. Among these, the Mid-Journey and Dalle samples were collected from social platforms.
\textbf{UClipiffusion~\cite{ojha2023towards}} is another diffusion dataset, collected from UClip~\cite{ojha2023towards}, encompassing four different diffusion models with varying configurations.
\textbf{Microsoft Northwestern Witness (MNW)~\cite{MNW2025}} is a recent and diverse dataset encompassing 43 diffusion models, with 250 samples generated for each model. The dataset includes samples from wild scenes, faces, and real-world scenarios.
\vspace{-16pt}

\begin{figure}[!t]
    \centering
    \vspace{8pt}
    \begin{tabular}{cccccc}
        \includegraphics[width=2.2cm]{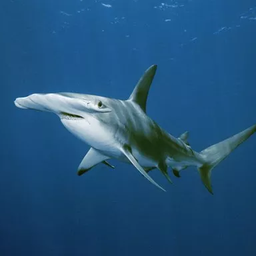} &
        \includegraphics[width=2.2cm]{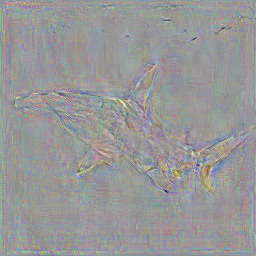} &
        \includegraphics[width=2.2cm]{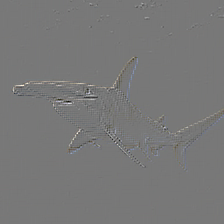} &
        \includegraphics[width=2.2cm]{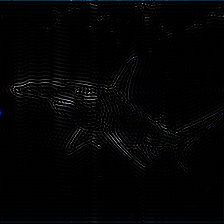} &
        \includegraphics[width=2.2cm]{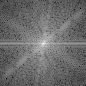} \\
        (a) & (b) & (c) & (d) & (e) \\
    \end{tabular}
    \vspace{1pt}
    \caption{Intermediate representation: (a) Original, (b) LGrad (gradient)~\cite{tan2023learning}, (c) NPR~\cite{tan2024rethinking}, (d) FreqNet (high-frequency)~\cite{tan2024frequency}, and (e) UpConv(spectral)~\cite{durall2020watch} .}
    \label{fig:intermediate}
    \vspace{-20pt}
\end{figure}

\subsubsection{Other Generative Datasets}\vspace{-8pt}
The \textbf{ForenSynthsCh~\cite{wang2020cnn}} dataset contains AI-generated images created using low-level vision and perceptual loss techniques, which are very challenging and often overlooked by most SoTA methods, as they worked very poorly.\vspace{-19pt}

\subsubsection{Detection Methods}\vspace{-8pt}
This section introduces the SoTA forensic methods used in our analysis, as summarized in Table~\ref{tab:methods}, and presents their intermediate representations in Figure~\ref{fig:intermediate} to better illustrate the underlying concepts.
\vspace{-18pt}

\subsubsection{Scratch Trained Models}\vspace{-8pt}
We selected four scratch-trained models~\cite{durall2020watch, tan2023learning, tan2024rethinking,tan2024frequency} to evaluate the proposed benchmark, each representing a distinct design in AI-generated image detection. UpConv~\cite{durall2020watch} is a widely recognized approach that exploits spectral analysis to identify upsampling artifacts, effectively capturing intrinsic properties of both GAN- and diffusion-generated content. LGrad~\cite{tan2023learning} is another influential method, which leverages a pretrained generative model to extract gradient-based features, thereby capturing subtle textural and structural cues associated with deepfakes. The nearest pixel relationship (NPR)~\cite{tan2024rethinking} method takes a spatial perspective, focusing on the correlations among neighboring pixels to uncover artifacts introduced during the upsampling process. Finally, FreqNet~\cite{tan2024frequency} represents a recent advancement in frequency-domain approaches, offering an end-to-end frequency-aware architecture capable of identifying nuanced spectral inconsistencies present in AI-generated images.
\vspace{-19pt}
\subsubsection{Frozen Models}\vspace{-8pt}
Similar to scratch-trained models, we selected two frozen-based models~\cite{ojha2023towards, cozzolino2024raising}. These two methods used CLIP as a frozen model to extract features for deepfake detection. First, universal deepfake detection using CLIP (UClip)~\cite{ojha2023towards}, in which the authors adopted a pretrained CLIP model for AI-generated image detection without training CLIP. RClip~\cite{cozzolino2024raising} investigated the effectiveness of sample size in generalizing detection with CLIP features, showing that even 0.01K samples are sufficient to detect deepfake artifacts.
\vspace{-18pt}
\subsubsection{Fine-Tuned Models}\vspace{-8pt}
This category of methods~\cite{BiometricThreat} fine-tunes pretrained models on AI-generated datasets to improve generalization. Examples include CNND~\cite{wang2020cnn}, RINE~\cite{koutlis2024leveraging}, FatF~\cite{liu2024forgery}, and C2PClip~\cite{c2pclip2025} . CNND~\cite{wang2020cnn} was the first to introduce a large-scale deepfake dataset, using a pretrained ResNet (trained on ImageNet) fine-tuned on this dataset. RINE~\cite{koutlis2024leveraging} employs a frozen CLIP encoder with a trainable importance estimator to select key features for AI-generated image detection. Similarly, FatFormer integrates a forgery-aware adapter to capture frequency cues, while C2PClip~\cite{c2pclip2025}  injects category-common prompts to enhance generalization.
\vspace{-20pt}
\section{{Results}}\vspace{-10pt}
This section presents the experimental setups, performance evaluations and comparisons, and explainability analyses conducted in the proposed empirical study.
\vspace{-18pt}
\label{sec:results}
\subsection{Experimental Setting}\vspace{-8pt}
We configured our pipeline to evaluate all selected methods under identical environmental settings. We run all the experiments on a Linux 24.04 operating system with eight NVIDIA RTX 6000 Ada Generation GPUs (49 GB of memory on each GPU). For each method, we adopted the preprocessing, including load size, cropping, and normalization, reported in the original papers. We reported ACC, AP, AUC, and EER for a fair assessment of the methods. 
\vspace{-28pt}
\subsection{Performance Evaluation and Comparisons}
\begin{figure*}[!t]%
\centering    
{\includegraphics[width=1\linewidth]{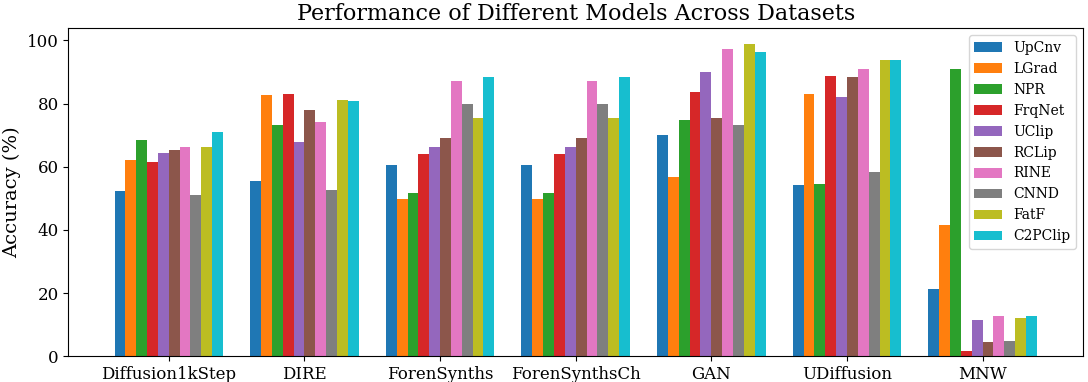}}
    \vspace{0.1pt}
    \caption{Summary of all forensic methods on all benchmark datasets.}
    \label{fig:summary}
    \vspace{-22pt}
\end{figure*}

\vspace{-8pt}
We extensively evaluated the performance of ten forensic methods on seven benchmark datasets, as reported in Tables~\ref{tab:diffusion1kstep}-\ref{tab:mnw}. For the CNND~\cite{wang2020cnn} dataset, we split it into two categories because most methods tend to ignore the ForenSynthsCh~\cite{wang2020cnn} segments. This is because many SoTA methods fail to generalize on this dataset, resulting in poor performance.\\
Across most datasets, UpConv~\cite{durall2020watch} underperforms, while C2PClip~\cite{c2pclip2025} consistently achieves the highest accuracy, demonstrating strong generalization. All methods struggle on Diffusion1kStep~\cite{tan2024rethinking}, whereas UDiffusion and GAN-based datasets are easier to detect, with several models exceeding 90\% accuracy. The best performance on Diffusion1kStep~\cite{tan2024rethinking} is achieved by C2PClip~\cite{c2pclip2025} (ACC/AP of 70.9\%/91.0\%), whereas the lowest results are reported by CNND~\cite{wang2020cnn} (ACC/AP of 51.1\%/54.7\%). Methods like LGrad~\cite{tan2023learning}, RCLip~\cite{cozzolino2024raising}, and CNND~\cite{wang2020cnn} show intermediate performance across all datasets, as depicted in Figure~\ref{fig:summary}. \\
Among all methods, NPR~\cite{tan2024rethinking} achieves the highest average accuracy (91.1\%) on the MNW~\cite{MNW2025} dataset, whereas most other models perform substantially worse and face the difficulty of generalizing across diverse generative sources. Notably, FreqNet~\cite{tan2024frequency} drops to only 1.6\% accuracy, despite its strong performance on other datasets, which highlights the challenges of adapting to certain real-world or unseen data distributions.
\vspace{-10pt}

\begin{table*}[!t]
\centering
\small
\renewcommand{\arraystretch}{1.1}
\setlength{\tabcolsep}{5pt}
\caption{Performance evaluation on Diffusion1kStep~\cite{tan2024rethinking} datasets (ACC/AP).}
\vspace{4pt}
\begin{adjustbox}{width=1\linewidth}
\begin{tabular}{lcccccccccc}
\hline
{\multirow{2}{*}{\textbf{Dataset}}} & \multicolumn{4}{c}{\textbf{Scratch Models}} & \multicolumn{2}{c}{\textbf{Frozen Models}} & \multicolumn{4}{c}{\textbf{Fine-Tuned Models}} \\
\cmidrule(lr){2-5} \cmidrule(lr){6-7} \cmidrule(lr){8-11}
& \textbf{UpConv~\cite{durall2020watch}} & \textbf{LGrad~\cite{tan2023learning}} & \textbf{NPR~\cite{tan2024rethinking}} & \textbf{FreqNet~\cite{tan2024frequency}} & \textbf{UClip~\cite{ojha2023towards}} & \textbf{RCLip~\cite{cozzolino2024raising}} & \textbf{RINE~\cite{koutlis2024leveraging}} & \textbf{CNND~\cite{wang2020cnn}} & \textbf{FatF~\cite{liu2024forgery}} & \textbf{C2PClip~\cite{c2pclip2025} } \\
\hline
Dalle & 47.9/46.5 & 76.6/87.1 & 69.5/86.4 & 51.3/58.9 & 53.7/69.3 & 76.8/81.1 & 60.5/86.2 & 52.8/53.4 & 68.8/93.2 & 66.4/93.1 \\
Ddpm & 49.9/49.8 & 59.8/80.3 & 70.8/81.9 & 69.4/86.6 & 72.2/84.4 & 65.6/74.1 & 68.6/85.1 & 50.2/59.0 & 59.1/77.9 & 72.0/81.9 \\
Guided-diffusion & 57.6/68.3 & 68.5/74.8 & 64.3/77.6 & 80.3/90.2 & 77.5/94.5 & 70.0/80.1 & 82.2/97.7 & 56.4/67.7 & 81.8/95.7 & 74.4/94.6 \\
Improved-diffusion & 53.8/62.8 & 42.3/43.9 & 68.7/87.0 & 52.9/60.5 & 69.2/90.8 & 51.3/50.1 & 66.6/92.8 & 47.3/51.4 & 59.4/72.5 & 75.2/92.6 \\
Midjourney & 51.7/53.4 & 64.1/71.4 & 68.8/88.2 & 53.4/61.7 & 49.9/48.5 & 62.7/65.3 & 53.3/67.1 & 49.0/42.1 & 62.7/85.4 & 66.8/93.1 \\
\hline
\textbf{Avg.} & \textbf{52.2/56.1} & \textbf{62.3/71.5} & \textbf{68.4/84.2} & \textbf{61.5/71.6} & \textbf{64.5/77.5} & \textbf{65.3/70.2} & \textbf{66.2/85.8} & \textbf{51.1/54.7} & \textbf{66.4/84.9} & \textbf{70.9/91.0} \\
\hline
\end{tabular}
\end{adjustbox}
\label{tab:diffusion1kstep}
\vspace{-19pt}
\end{table*}

\begin{table*}[!t]
\centering
\small
\renewcommand{\arraystretch}{1.1}
\setlength{\tabcolsep}{5pt}
\caption{Performance evaluation on DIRE~\cite{wang2023dire} datasets (ACC/AP).}
\vspace{4pt}
\begin{adjustbox}{width=1\linewidth}
\begin{tabular}{lcccccccccc}
\hline
{\multirow{2}{*}{\textbf{Dataset}}} & \multicolumn{4}{c}{\textbf{Scratch Models}} & \multicolumn{2}{c}{\textbf{Frozen Models}} & \multicolumn{4}{c}{\textbf{Fine-Tuned Models}} \\
\cmidrule(lr){2-5} \cmidrule(lr){6-7} \cmidrule(lr){8-11}
& \textbf{UpConv~\cite{durall2020watch}} & \textbf{LGrad~\cite{tan2023learning}} & \textbf{NPR~\cite{tan2024rethinking}} & \textbf{FreqNet~\cite{tan2024frequency}} & \textbf{UClip~\cite{ojha2023towards}} & \textbf{RCLip~\cite{cozzolino2024raising}} & \textbf{RINE~\cite{koutlis2024leveraging}} & \textbf{CNND~\cite{wang2020cnn}} & \textbf{FatF~\cite{liu2024forgery}} & \textbf{C2PClip~\cite{c2pclip2025}} \\
\hline
Adm & 56.1/65.0 & 83.8/94.3 & 68.8/80.8 & 66.7/85.2 & 67.9/86.3 & 81.6/96.4 & 69.7/92.5 & 58.0/74.8 & 70.7/93.7 & 68.8/95.3 \\
Ddpm & 55.1/33.6 & 81.2/92.4 & 67.2/97.2 & 90.3/99.1 & 80.7/96.4 & 72.1/69.2 & 80.7/96.8 & 62.9/64.3 & 67.2/78.9 & 73.5/76.2 \\
Iddpm & 46.9/46.1 & 63.3/84.9 & 71.8/94.3 & 60.1/92.9 & 73.4/96.7 & 69.7/82.2 & 75.2/97.9 & 50.4/74.9 & 69.3/96.3 & 80.7/94.9 \\
Ldm & 63.5/67.2 & 98.7/99.9 & 74.0/99.6 & 97.5/100.0 & 50.7/86.1 & 95.6/100.0 & 56.6/98.1 & 53.0/75.8 & 97.2/100.0 & 97.2/99.7 \\
Pndm & 52.4/53.6 & 67.8/94.2 & 73.2/85.9 & 85.0/99.3 & 86.2/99.1 & 95.5/99.8 & 83.8/99.0 & 50.9/76.6 & 99.2/100.0 & 84.2/97.2 \\
Sdv1 & 42.0/74.2 & 83.2/97.5 & 82.4/94.9 & 93.8/99.6 & 52.8/90.8 & 68.0/96.8 & 78.0/98.8 & 39.1/78.0 & 61.6/97.0 & 78.9/99.2 \\
Sdv2 & 61.6/67.1 & 96.7/99.8 & 74.0/98.9 & 70.7/96.5 & 53.3/85.0 & 46.2/36.2 & 57.4/89.9 & 52.2/72.9 & 84.4/98.7 & 66.7/94.8 \\
Vqdiffusion & 65.3/70.4 & 86.1/99.0 & 74.0/99.6 & 99.9/100.0 & 77.8/99.0 & 95.6/100.0 & 91.4/99.9 & 53.9/84.7 & 100.0/100.0 & 95.8/99.7 \\
\hline
\textbf{Avg.} & \textbf{55.4/59.6} & \textbf{82.6/95.3} & \textbf{73.2/93.9} & \textbf{83.0/96.6} & \textbf{67.9/92.4} & \textbf{78.0/85.1} & \textbf{74.1/96.6} & \textbf{52.6/75.2} & \textbf{81.2/95.6} & \textbf{80.7/94.6} \\
\hline
\end{tabular}
\end{adjustbox}
\label{tab:dire}
\vspace{-8pt}
\end{table*}

\begin{table*}[!t]
\centering
\small
\renewcommand{\arraystretch}{1.1}
\setlength{\tabcolsep}{5pt}
\caption{Performance evaluation on ForenSynths~\cite{wang2020cnn} datasets (ACC/AP).}
\vspace{4pt}
\begin{adjustbox}{width=1\linewidth}
\begin{tabular}{lcccccccccc}
\hline
{\multirow{2}{*}{\textbf{Dataset}}} & \multicolumn{4}{c}{\textbf{Scratch Models}} & \multicolumn{2}{c}{\textbf{Frozen Models}} & \multicolumn{4}{c}{\textbf{Fine-Tuned Models}} \\
\cmidrule(lr){2-5} \cmidrule(lr){6-7} \cmidrule(lr){8-11}
& \textbf{UpConv~\cite{durall2020watch}} & \textbf{LGrad~\cite{tan2023learning}} & \textbf{NPR~\cite{tan2024rethinking}} & \textbf{FreqNet~\cite{tan2024frequency}} & \textbf{UClip~\cite{ojha2023towards}} & \textbf{RCLip~\cite{cozzolino2024raising}} & \textbf{RINE~\cite{koutlis2024leveraging}} & \textbf{CNND~\cite{wang2020cnn}} & \textbf{FatF~\cite{liu2024forgery}} & \textbf{C2PClip~\cite{c2pclip2025}} \\
\hline
Biggan & 67.3/81.9 & 74.5/78.3 & 58.4/65.2 & 91.2/96.2 & 95.1/99.3 & 80.4/95.6 & 99.6/99.9 & 70.2/84.5 & 99.5/100.0 & 99.1/100.0 \\
Cyclegan & 69.7/79.3 & 80.1/88.3 & 73.8/71.3 & 95.5/99.6 & 98.3/99.8 & 93.5/99.5 & 99.3/100.0 & 85.2/93.5 & 99.4/100.0 & 97.3/100.0 \\
Gaugan & 59.6/74.1 & 68.8/73.4 & 53.5/49.7 & 92.9/98.4 & 99.5/100.0 & 91.8/97.9 & 99.8/100.0 & 78.9/89.5 & 99.4/100.0 & 99.2/100.0 \\
Progan & 53.1/78.8 & 98.8/99.9 & 58.1/71.7 & 99.6/100.0 & 99.8/100.0 & 84.0/99.7 & 100.0/100.0 & 100.0/100.0 & 99.9/100.0 & 100.0/100.0 \\
Stargan & 92.8/100.0 & 95.7/99.8 & 63.5/99.0 & 84.3/99.3 & 95.7/99.4 & 61.4/98.8 & 99.5/100.0 & 91.7/98.1 & 99.7/100.0 & 99.6/100.0 \\
Stylegan & 60.1/74.7 & 92.6/99.3 & 65.4/84.6 & 91.2/99.8 & 84.9/97.6 & 84.9/94.0 & 88.9/99.4 & 87.1/99.6 & 97.1/99.8 & 96.4/99.5 \\
Stylegan2 & 53.8/68.6 & 93.6/99.2 & 61.7/74.8 & 87.3/99.5 & 75.0/97.9 & 80.8/90.2 & 94.5/100.0 & 84.4/99.1 & 98.8/99.9 & 95.6/99.9 \\
Deepfake & 53.6/53.5 & 58.9/81.8 & 49.9/52.9 & 92.2/97.3 & 68.6/81.8 & 53.3/72.8 & 80.6/97.9 & 53.5/89.0 & 93.3/98.0 & 93.8/98.6 \\
\hline
\textbf{Avg.} & \textbf{63.8/76.4} & \textbf{82.9/90.0} & \textbf{60.5/71.2} & \textbf{91.8/98.8} & \textbf{89.6/97.0} & \textbf{78.7/93.6} & \textbf{95.3/99.7} & \textbf{81.4/94.2} & \textbf{98.4/99.7} & \textbf{97.6/99.7} \\
\hline
\end{tabular}
\end{adjustbox}
\label{tab:forensynths}
\vspace{-12pt}
\end{table*}

\begin{table*}[!t]
\centering
\small
\renewcommand{\arraystretch}{1.1}
\setlength{\tabcolsep}{5pt}
\caption{Performance evaluation on ForenSynthsCh~\cite{wang2020cnn} datasets (ACC/AP).}
\vspace{4pt}
\begin{adjustbox}{width=1\linewidth}
\begin{tabular}{lcccccccccc}
\hline
{\multirow{2}{*}{\textbf{Dataset}}} & \multicolumn{4}{c}{\textbf{Scratch Models}} & \multicolumn{2}{c}{\textbf{Frozen Models}} & \multicolumn{4}{c}{\textbf{Fine-Tuned Models}} \\
\cmidrule(lr){2-5} \cmidrule(lr){6-7} \cmidrule(lr){8-11}
& \textbf{UpConv~\cite{durall2020watch}} & \textbf{LGrad~\cite{tan2023learning}} & \textbf{NPR~\cite{tan2024rethinking}} & \textbf{FreqNet~\cite{tan2024frequency}} & \textbf{UClip~\cite{ojha2023towards}} & \textbf{RCLip~\cite{cozzolino2024raising}} & \textbf{RINE~\cite{koutlis2024leveraging}} & \textbf{CNND~\cite{wang2020cnn}} & \textbf{FatF~\cite{liu2024forgery}} & \textbf{C2PClip~\cite{c2pclip2025}} \\
\hline
CRN & 52.5/60.1 & 51.2/64.7 & 48.8/45.5 & 53.7/74.8 & 56.6/96.6 & 61.3/83.1 & 89.3/97.3 & 86.3/98.2 & 69.5/99.8 & 93.3/99.9 \\
IMLE & 51.6/62.5 & 51.2/70.9 & 48.8/50.7 & 53.7/69.9 & 69.1/98.6 & 66.1/83.2 & 90.7/99.7 & 86.2/98.4 & 69.5/99.9 & 93.3/99.9 \\
SAN & 50.5/48.0 & 42.0/41.3 & 58.7/68.4 & 89.3/93.2 & 56.6/78.8 & 76.5/88.0 & 68.3/94.9 & 50.5/70.4 & 68.0/81.2 & 64.4/84.6 \\
SITD & 85.0/97.1 & 47.2/39.1 & 51.7/53.0 & 72.8/72.1 & 62.2/63.8 & 70.6/91.2 & 90.6/97.2 & 90.3/97.2 & 81.4/97.9 & 95.6/98.9 \\
WFR & 64.1/84.0 & 57.8/58.9 & 51.0/49.4 & 50.9/96.7 & 87.2/97.3 & 71.4/90.3 & 97.0/99.5 & 86.8/94.8 & 88.1/98.5 & 94.8/99.5 \\
\hline
\textbf{Avg.} & \textbf{60.7/70.3} & \textbf{49.9/55.0} & \textbf{51.8/53.4} & \textbf{64.1/81.3} & \textbf{66.3/87.0} & \textbf{69.1/87.2} & \textbf{87.2/97.7} & \textbf{80.0/91.8} & \textbf{75.3/95.5} & \textbf{88.3/96.6} \\
\hline
\end{tabular}
\end{adjustbox}
\label{tab:forensynthsch}
\vspace{-12pt}
\end{table*}

\begin{table*}[!t]
\centering
\small
\renewcommand{\arraystretch}{1.1}
\setlength{\tabcolsep}{5pt}
\caption{Performance evaluation on GAN~\cite{tan2024rethinking} datasets (ACC/AP).}
\vspace{4pt}
\begin{adjustbox}{width=1\linewidth}
\begin{tabular}{lcccccccccc}
\hline
{\multirow{2}{*}{\textbf{Dataset}}} & \multicolumn{4}{c}{\textbf{Scratch Models}} & \multicolumn{2}{c}{\textbf{Frozen Models}} & \multicolumn{4}{c}{\textbf{Fine-Tuned Models}} \\
\cmidrule(lr){2-5} \cmidrule(lr){6-7} \cmidrule(lr){8-11}
& \textbf{UpConv~\cite{durall2020watch}} & \textbf{LGrad~\cite{tan2023learning}} & \textbf{NPR~\cite{tan2024rethinking}} & \textbf{FreqNet~\cite{tan2024frequency}} & \textbf{UClip~\cite{ojha2023towards}} & \textbf{RCLip~\cite{cozzolino2024raising}} & \textbf{RINE~\cite{koutlis2024leveraging}} & \textbf{CNND~\cite{wang2020cnn}} & \textbf{FatF~\cite{liu2024forgery}} & \textbf{C2PClip~\cite{c2pclip2025}} \\
\hline
Attgan & 48.5/41.9 & 53.1/76.6 & 86.4/98.0 & 90.3/98.5 & 90.8/97.0 & 81.3/94.9 & 99.2/100.0 & 65.8/91.4 & 99.3/100.0 & 90.4/99.8 \\
Began & 48.9/47.9 & 51.0/70.4 & 55.2/78.7 & 65.4/99.3 & 89.3/96.3 & 99.9/100.0 & 97.9/99.9 & 69.7/91.9 & 99.9/100.0 & 94.8/100.0 \\
Cramergan & 73.5/84.4 & 50.9/59.1 & 73.4/92.7 & 99.6/100.0 & 90.7/99.3 & 68.0/90.0 & 97.0/99.9 & 91.9/99.1 & 98.4/100.0 & 98.4/100.0 \\
Infomaxgan & 42.2/42.2 & 53.9/82.1 & 74.4/92.6 & 63.2/95.0 & 88.5/96.9 & 68.0/90.2 & 96.5/99.6 & 62.5/86.7 & 98.4/100.0 & 98.4/100.0 \\
Mmdgan & 76.1/87.0 & 51.1/66.5 & 74.0/93.5 & 98.0/99.9 & 90.6/99.2 & 68.0/90.1 & 97.0/99.9 & 86.4/98.2 & 98.4/100.0 & 98.4/100.0 \\
Relgan & 93.7/98.2 & 74.5/95.6 & 88.1/99.9 & 99.9/100.0 & 93.4/98.0 & 80.1/98.8 & 99.4/100.0 & 88.8/98.9 & 99.5/100.0 & 92.0/99.8 \\
S3gan & 96.5/99.6 & 73.3/75.9 & 73.2/82.7 & 88.6/94.1 & 94.1/98.8 & 85.1/99.0 & 98.6/99.9 & 69.0/80.7 & 99.0/100.0 & 99.0/100.0 \\
Sngan & 65.5/73.3 & 52.3/82.5 & 57.8/64.4 & 51.2/84.7 & 88.6/96.8 & 67.9/81.7 & 96.7/99.7 & 60.8/86.6 & 98.3/99.9 & 98.4/99.9 \\
Stgan & 85.7/95.9 & 50.5/75.7 & 91.4/99.1 & 98.0/100.0 & 82.8/91.6 & 61.5/89.8 & 93.7/99.1 & 65.2/96.5 & 98.8/99.8 & 97.6/99.6 \\
\hline
\textbf{Avg.} & \textbf{70.1/74.5} & \textbf{56.7/76.1} & \textbf{74.9/89.1} & \textbf{83.8/96.8} & \textbf{89.9/97.1} & \textbf{75.5/92.7} & \textbf{97.3/99.8} & \textbf{73.3/92.2} & \textbf{98.9/100.0} & \textbf{96.4/99.9} \\
\hline
\end{tabular}
\end{adjustbox}
\label{tab:selfsynthesisgan}
\vspace{-12pt}
\end{table*}

\begin{table*}[!t]
\centering
\small
\renewcommand{\arraystretch}{1.1}
\setlength{\tabcolsep}{5pt}
\caption{Performance evaluation on UClipiffusion~\cite{ojha2023towards} datasets (ACC/AP).}
\vspace{4pt}
\begin{adjustbox}{width=1\linewidth}
\begin{tabular}{lcccccccccc}
\hline
{\multirow{2}{*}{\textbf{Dataset}}} & \multicolumn{4}{c}{\textbf{Scratch Models}} & \multicolumn{2}{c}{\textbf{Frozen Models}} & \multicolumn{4}{c}{\textbf{Fine-Tuned Models}} \\
\cmidrule(lr){2-5} \cmidrule(lr){6-7} \cmidrule(lr){8-11}
& \textbf{UpConv~\cite{durall2020watch}} & \textbf{LGrad~\cite{tan2023learning}} & \textbf{NPR~\cite{tan2024rethinking}} & \textbf{FreqNet~\cite{tan2024frequency}} & \textbf{UClip~\cite{ojha2023towards}} & \textbf{RCLip~\cite{cozzolino2024raising}} & \textbf{RINE~\cite{koutlis2024leveraging}} & \textbf{CNND~\cite{wang2020cnn}} & \textbf{FatF~\cite{liu2024forgery}} & \textbf{C2PClip~\cite{c2pclip2025}} \\
\hline
Dalle & 55.1/65.5 & 83.5/92.4 & 53.8/69.5 & 97.7/99.5 & 87.5/97.7 & 89.2/99.5 & 95.0/99.5 & 56.1/71.3 & 98.7/99.8 & 98.6/99.9 \\
Glide\_50\_27 & 58.1/67.0 & 85.2/92.3 & 54.0/80.8 & 86.6/95.8 & 79.2/96.0 & 87.2/96.7 & 92.6/99.5 & 62.7/84.6 & 94.6/99.5 & 95.2/99.8 \\
Glide\_100\_10 & 59.7/69.1 & 83.7/91.5 & 54.1/81.0 & 88.4/96.2 & 78.0/95.5 & 87.9/97.0 & 90.7/99.2 & 61.0/82.0 & 94.2/99.3 & 96.1/99.8 \\
Glide\_100\_27 & 54.5/60.7 & 81.5/89.2 & 53.9/80.0 & 84.7/95.4 & 78.6/95.8 & 87.8/97.0 & 88.9/99.1 & 60.4/80.5 & 94.3/99.3 & 95.2/99.7 \\
Guided & 57.5/68.7 & 70.2/75.1 & 58.8/67.3 & 62.4/67.2 & 70.0/88.3 & 85.6/96.6 & 76.1/96.6 & 62.0/77.7 & 76.0/91.9 & 69.1/94.1 \\
Ldm\_100 & 49.5/54.9 & 86.4/93.7 & 54.4/82.7 & 97.0/99.9 & 95.2/99.3 & 89.5/99.9 & 98.7/99.9 & 55.1/72.5 & 98.6/99.9 & 99.3/100.0 \\
Ldm\_200\_cfg & 51.3/56.7 & 88.2/95.4 & 54.3/82.9 & 96.9/99.8 & 74.2/93.2 & 89.3/99.7 & 88.2/98.7 & 55.2/73.0 & 94.8/99.2 & 97.2/99.8 \\
Ldm\_200 & 49.0/54.2 & 86.1/93.7 & 54.4/82.6 & 96.9/99.8 & 94.5/99.4 & 89.5/99.9 & 98.3/99.9 & 53.9/71.1 & 98.6/99.8 & 99.2/100.0 \\
\hline
\textbf{Avg.} & \textbf{54.3/62.1} & \textbf{83.1/90.4} & \textbf{54.7/78.4} & \textbf{88.8/94.2} & \textbf{82.2/95.7} & \textbf{88.3/98.3} & \textbf{91.1/99.0} & \textbf{58.3/76.6} & \textbf{93.7/98.6} & \textbf{93.8/99.1} \\
\hline
\end{tabular}
\end{adjustbox}
\label{tab:UClipiffusion}
\vspace{-19pt}
\end{table*}

\begin{table*}[!t]
\centering
\small
\renewcommand{\arraystretch}{1.1}
\setlength{\tabcolsep}{5pt}
\caption{Performance evaluation on MNW~\cite{MNW2025} datasets (ACC/AP).}
\vspace{4pt}
\begin{adjustbox}{width=1\linewidth}
\begin{tabular}{lcccccccccc}
\hline
{\multirow{2}{*}{\textbf{Dataset}}} & \multicolumn{4}{c}{\textbf{Scratch Models}} & \multicolumn{2}{c}{\textbf{Frozen Models}} & \multicolumn{4}{c}{\textbf{Fine-Tuned Models}} \\
\cmidrule(lr){2-5} \cmidrule(lr){6-7} \cmidrule(lr){8-11}
& \textbf{UpConv~\cite{durall2020watch}} & \textbf{LGrad~\cite{tan2023learning}} & \textbf{NPR~\cite{tan2024rethinking}} & \textbf{FreqNet~\cite{tan2024frequency}} & \textbf{UClip~\cite{ojha2023towards}} & \textbf{RCLip~\cite{cozzolino2024raising}} & \textbf{RINE~\cite{koutlis2024leveraging}} & \textbf{CNND~\cite{wang2020cnn}} & \textbf{FatF~\cite{liu2024forgery}} & \textbf{C2PClip~\cite{c2pclip2025}} \\
\hline
Adobe & 41.6/-- & 56.5/-- & 97.6/-- & 1.2/-- & 28.9/-- & 4.7/-- & 38.5/-- & 3.1/-- & 16.7/-- & 31.3/-- \\
Adversarial\_images & 15.2/-- & 23.6/-- & 89.2/-- & 2.4/-- & 10.4/-- & 5.2/-- & 8.8/-- & 1.2/-- & 6.4/-- & 6.4/-- \\
Amazon\_titan\_v2 & 10.8/-- & 72.0/-- & 100.0/-- & 0.0/-- & 20.0/-- & 29.6/-- & 27.6/-- & 3.2/-- & 23.6/-- & 26.4/-- \\
Aura\_flow & 18.4/-- & 56.8/-- & 100.0/-- & 0.8/-- & 2.4/-- & 4.8/-- & 3.6/-- & 0.0/-- & 12.0/-- & 47.2/-- \\
Baidu & 10.4/-- & 13.2/-- & 85.2/-- & 0.8/-- & 10.4/-- & 1.2/-- & 6.8/-- & 0.8/-- & 0.4/-- & 4.4/-- \\
Bytedance & 31.6/-- & 49.6/-- & 99.2/-- & 0.0/-- & 0.0/-- & 1.2/-- & 0.4/-- & 0.4/-- & 0.0/-- & 0.0/-- \\
Civitai\_v6 & 3.6/-- & 80.0/-- & 98.0/-- & 0.0/-- & 4.0/-- & 12.0/-- & 4.8/-- & 0.8/-- & 21.6/-- & 28.0/-- \\
Flux & 12.8/-- & 35.2/-- & 88.0/-- & 15.1/-- & 3.7/-- & 2.1/-- & 3.1/-- & 2.4/-- & 2.9/-- & 3.1/-- \\
Google & 5.8/-- & 13.8/-- & 70.8/-- & 2.2/-- & 4.8/-- & 1.4/-- & 1.0/-- & 2.0/-- & 0.0/-- & 0.2/-- \\
Hunyuandit & 32.0/-- & 1.6/-- & 94.0/-- & 0.0/-- & 7.2/-- & 1.2/-- & 4.4/-- & 0.8/-- & 0.4/-- & 11.2/-- \\
Hypersd & 19.6/-- & 3.6/-- & 68.0/-- & 0.6/-- & 3.2/-- & 0.0/-- & 0.8/-- & 1.4/-- & 0.0/-- & 3.4/-- \\
Ideogram & 6.8/-- & 78.4/-- & 98.0/-- & 0.4/-- & 2.0/-- & 0.0/-- & 2.0/-- & 4.0/-- & 14.8/-- & 1.2/-- \\
Kandinsky & 17.2/-- & 1.6/-- & 88.4/-- & 1.2/-- & 11.2/-- & 0.0/-- & 4.0/-- & 0.0/-- & 0.0/-- & 6.0/-- \\
Krea\_1 & 6.0/-- & 90.4/-- & 98.8/-- & 0.0/-- & 4.8/-- & 0.0/-- & 5.6/-- & 6.0/-- & 7.2/-- & 1.2/-- \\
Kuaishou\_kolors & 8.0/-- & 2.8/-- & 85.2/-- & 1.6/-- & 3.2/-- & 0.4/-- & 0.8/-- & 0.4/-- & 0.0/-- & 9.6/-- \\
Luma\_photon & 97.6/-- & 84.0/-- & 99.2/-- & 1.2/-- & 26.8/-- & 5.2/-- & 45.2/-- & 14.4/-- & 41.6/-- & 16.8/-- \\
Lumina & 20.4/-- & 82.4/-- & 99.2/-- & 0.4/-- & 15.6/-- & 13.2/-- & 30.4/-- & 8.4/-- & 27.6/-- & 10.8/-- \\
Meta\_imagine & 4.0/-- & 15.2/-- & 94.0/-- & 1.2/-- & 14.8/-- & 2.8/-- & 12.0/-- & 5.6/-- & 0.0/-- & 12.4/-- \\
Midjourney & 30.4/-- & 30.8/-- & 78.9/-- & 0.8/-- & 5.7/-- & 0.0/-- & 8.8/-- & 10.9/-- & 6.1/-- & 7.3/-- \\
Nvidia\_sana & 47.6/-- & 22.4/-- & 95.6/-- & 1.2/-- & 60.4/-- & 30.4/-- & 60.4/-- & 0.4/-- & 49.6/-- & 54.4/-- \\
Openai & 7.3/-- & 37.9/-- & 86.5/-- & 2.4/-- & 20.3/-- & 2.0/-- & 20.7/-- & 8.0/-- & 2.9/-- & 14.5/-- \\
Pixart\_alpha\_xl & 32.0/-- & 3.6/-- & 82.0/-- & 0.8/-- & 2.0/-- & 0.8/-- & 0.8/-- & 1.2/-- & 0.0/-- & 12.0/-- \\
Playgroundai & 22.0/-- & 38.8/-- & 80.2/-- & 0.0/-- & 6.2/-- & 1.0/-- & 9.2/-- & 3.2/-- & 27.4/-- & 10.2/-- \\
Recraft\_v3 & 46.0/-- & 66.0/-- & 93.6/-- & 0.0/-- & 12.4/-- & 0.4/-- & 7.2/-- & 0.0/-- & 6.8/-- & 1.6/-- \\
Reve\_ai & 13.2/-- & 72.0/-- & 99.2/-- & 0.8/-- & 2.8/-- & 0.0/-- & 7.2/-- & 4.0/-- & 13.2/-- & 1.2/-- \\
Stable\_diffusion & 14.2/-- & 41.1/-- & 89.6/-- & 2.0/-- & 14.3/-- & 9.2/-- & 13.7/-- & 3.9/-- & 17.1/-- & 17.2/-- \\
Ultrapixel & 11.2/-- & 84.0/-- & 100.0/-- & 5.6/-- & 4.0/-- & 0.0/-- & 4.0/-- & 44.4/-- & 38.8/-- & 0.8/-- \\
Wuerstchen & 8.4/-- & 4.0/-- & 93.6/-- & 1.2/-- & 22.4/-- & 1.2/-- & 30.4/-- & 5.6/-- & 0.8/-- & 17.2/-- \\
\hline
\textbf{Avg.} & \textbf{21.2/--} & \textbf{41.5/--} & \textbf{91.1/--} & \textbf{1.6/--} & \textbf{11.6/--} & \textbf{4.6/--} & \textbf{12.9/--} & \textbf{4.9/--} & \textbf{12.1/--} & \textbf{12.7/--} \\
\hline
\end{tabular}
\end{adjustbox}
\label{tab:mnw}
\vspace{-12pt}
\end{table*}

\subsection{Explainability of Model Predictions}
\begin{figure}[!t]
    \centering
    \begin{tabular}{cccccc}
        \includegraphics[width=2.2cm]{figures/orig.png} &
        \includegraphics[width=2.2cm]{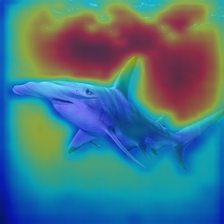} &
        \includegraphics[width=2.2cm]{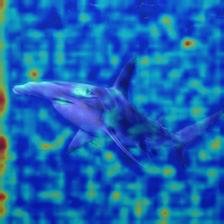} &
        \includegraphics[width=2.2cm]{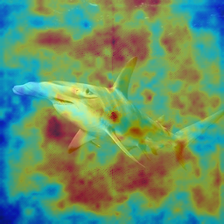} &
        \includegraphics[width=2.2cm]{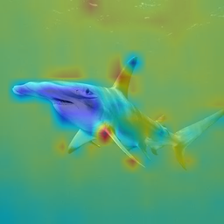} \\
        (a) & (b) & (c) & (d) & (e) \\
        \includegraphics[width=2.2cm]{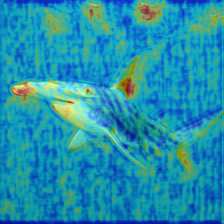} &
        \includegraphics[width=2.2cm]{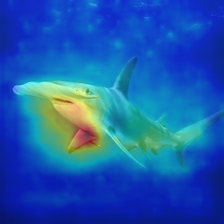} &
        \includegraphics[width=2.2cm]{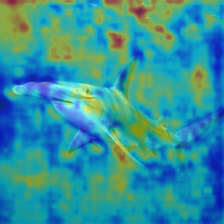} &
        \includegraphics[width=2.2cm]{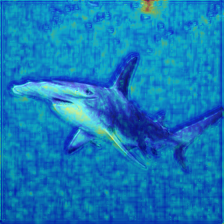} &
        \includegraphics[width=2.2cm]{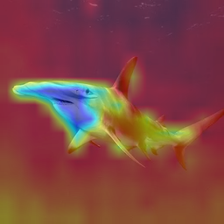} \\
        (f) & (g) & (h) & (i) & (j) \\
    \end{tabular}
    \vspace{2pt}
    \caption{GradCAM explanation: (a) Original, (b) LGrad~\cite{tan2023learning}, (c) NPR~\cite{tan2024rethinking}, (d) FreqNet~\cite{tan2024frequency}, (e) UClip~\cite{ojha2023towards}, (f) RClip~\cite{cozzolino2024raising}, (g) RINE~\cite{koutlis2024leveraging}, (h) CNND~\cite{wang2020cnn}, (i) FatF~\cite{liu2024forgery}, and (j) C2PClip~\cite{c2pclip2025}.}
    \label{fig:gradcam}
    \vspace{-16pt}
\end{figure}

\begin{figure}[!t]
    \centering
    \begin{tabular}{ccc}
        \includegraphics[width=3.9cm]{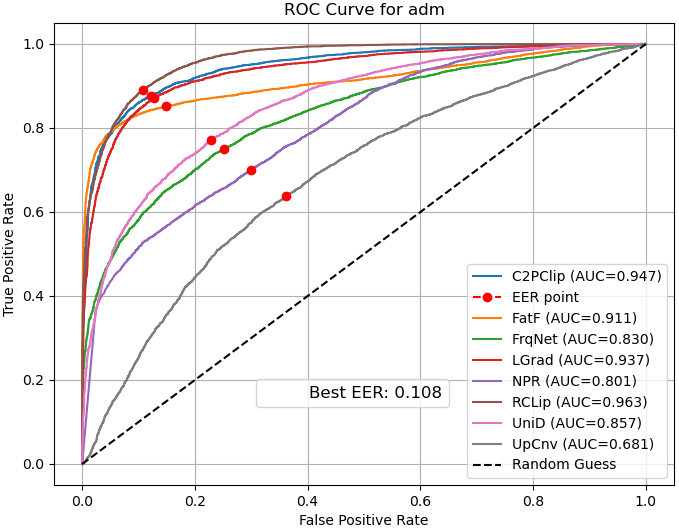} &
        \includegraphics[width=3.9cm]{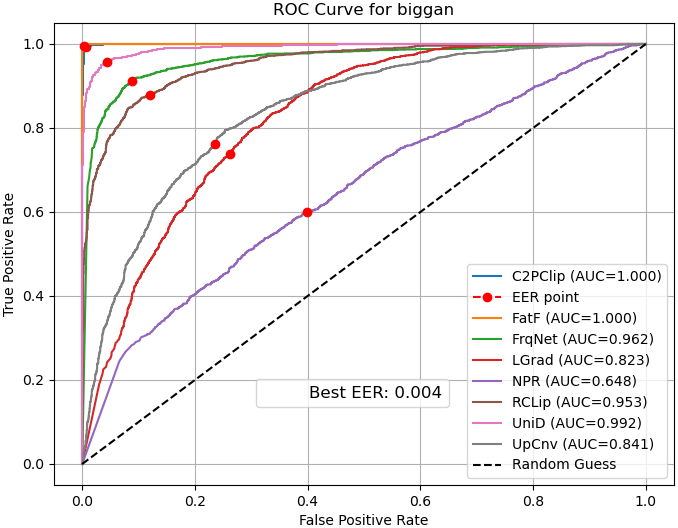} &
        \includegraphics[width=3.9cm]{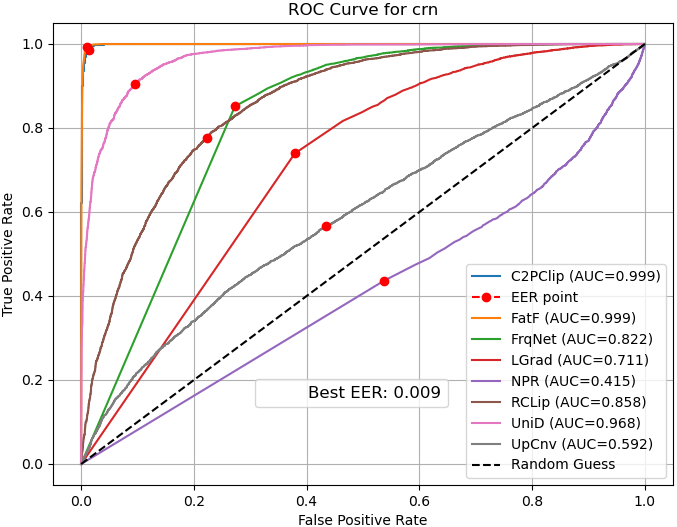} \\
        (a) & (b) & (c) \\  
        \includegraphics[width=3.9cm]{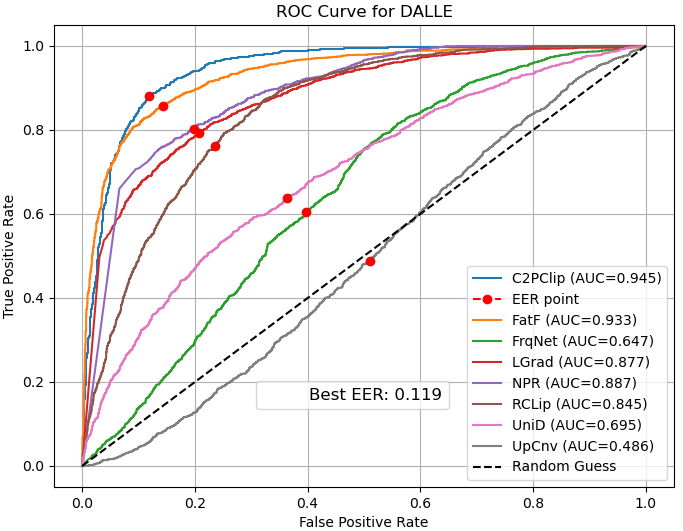} &
        \includegraphics[width=3.9cm]{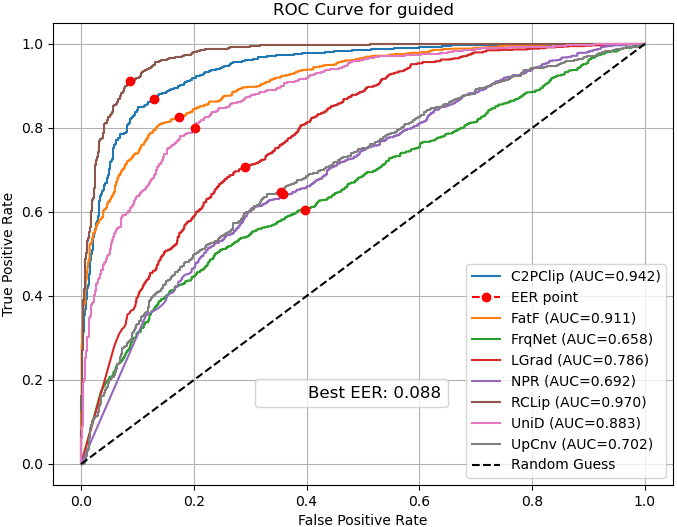} &
        \includegraphics[width=3.9cm]{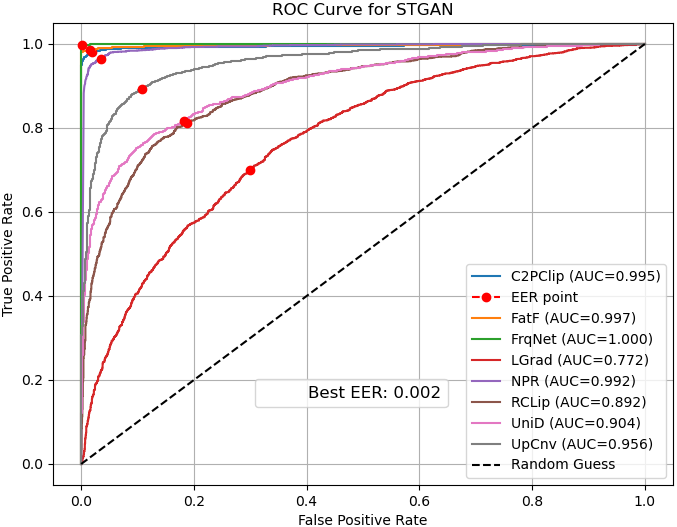} \\
        (d) & (e) & (f)
    \end{tabular}
    \vspace{2pt}
    \caption{ROC curve of each model corresponding to prediction confidence: (a) Adm, (b) BigGAN, (c) CRN, (d) Dalle, (e) Guided, and (f) StGAN.}
    \label{fig:roc_curve}
    \vspace{-19pt}
\end{figure}

\vspace{-8pt}
For a better explanation of model predictions, we visualized GradCAM, confidence, and ROC curves, as depicted in Figures~\ref{fig:gradcam}, \ref{fig:conf_curve}, and \ref{fig:roc_curve}. GradCAM highlights the regions that each model focuses on to distinguish between real and fake samples. As shown in Figure~\ref{fig:gradcam}, each method focuses on different regions to determine whether a sample is real. For example, LGrad~\cite{tan2023learning}, FreqNet~\cite{tan2024frequency}, and C2PClip~\cite{c2pclip2025} primarily target the background, while others attend to random regions when making their decisions.\\
In contrast, the confidence curve represents the prediction probabilities of each model for the real and fake classes to make it clear how confident a model is in predicting real as real and fake as fake. As shown in Figure~\ref{fig:conf_curve}, in most cases, NPR~\cite{tan2024rethinking} is biased towards the fake class, while UpConv~\cite{durall2020watch} tends to favor the real class. Similar to the confidence curve, the ROC curve illustrates the trade-off between the true positive rate and false positive rate across different thresholds to provide a comprehensive view of each model’s discriminative ability.
\vspace{-28pt}

\section{Discussions and Future Research Directions}\vspace{-8pt}
This section outlines the discussions and future research directions of our findings.\vspace{-18pt}
\subsection{Discussions}
\begin{figure}[!t]
    \centering
    \vspace{2pt}
    \begin{tabular}{ccc}
        \includegraphics[width=3.9cm]{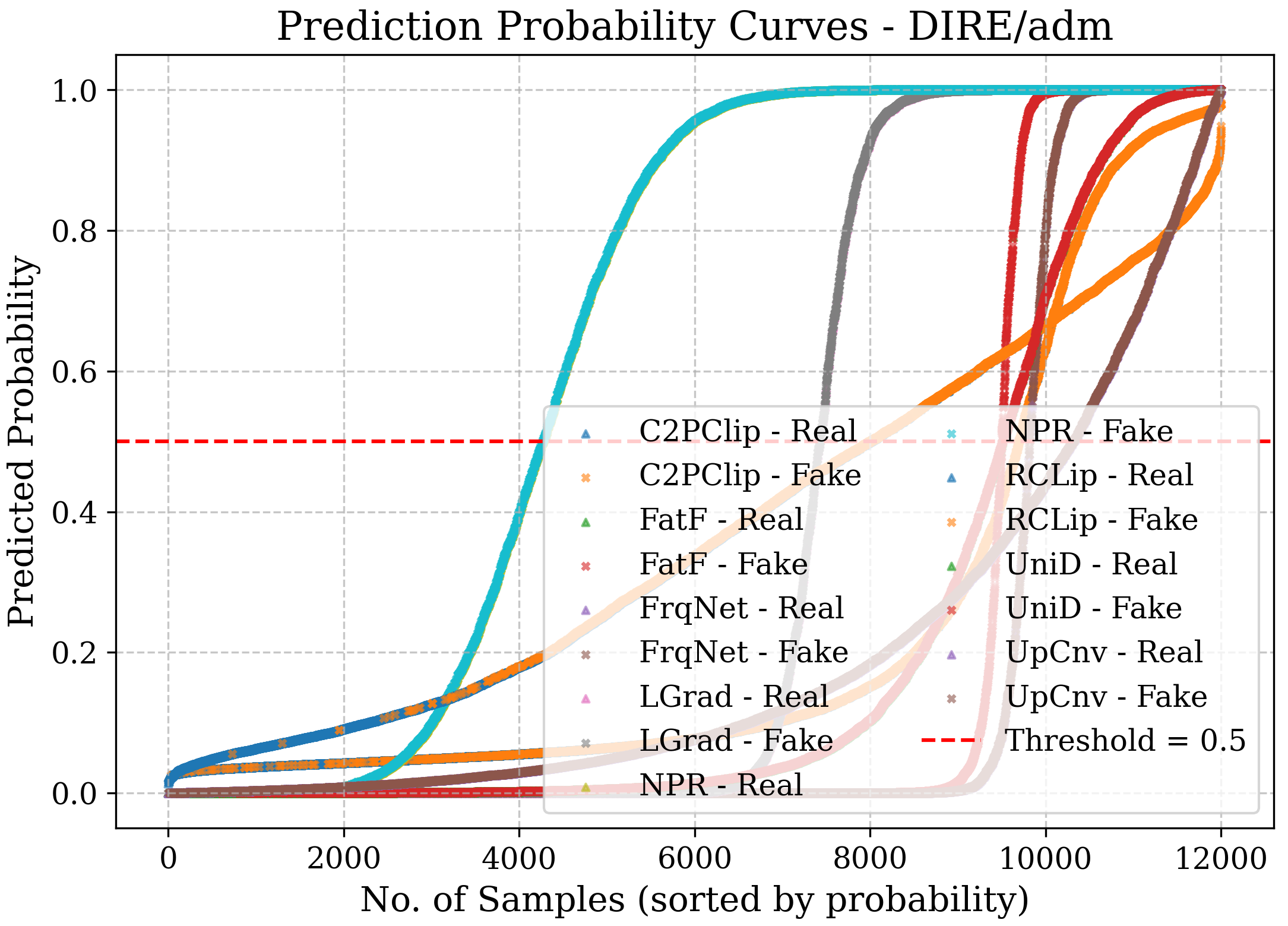} &
        \includegraphics[width=3.9cm]{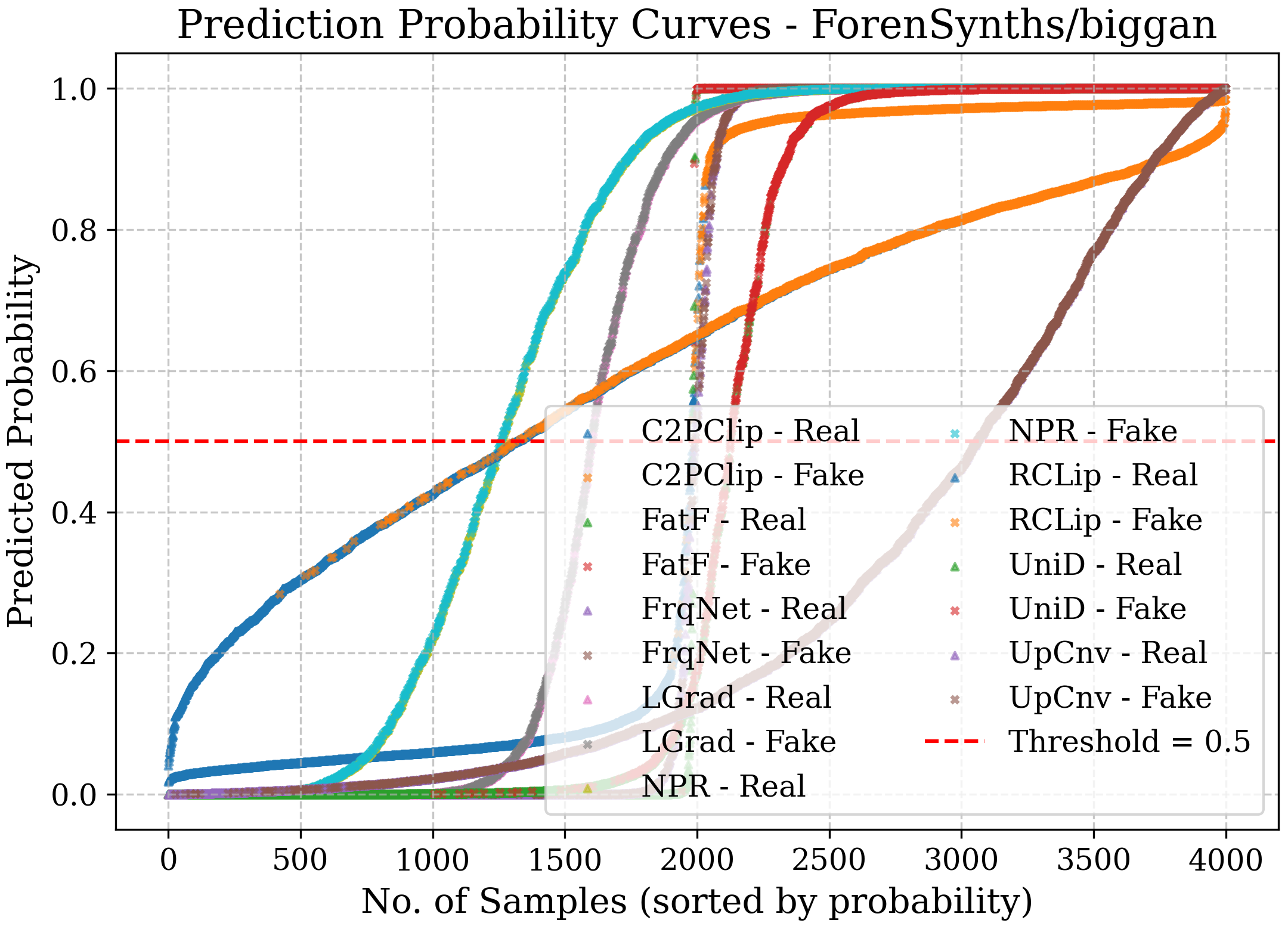} &
        \includegraphics[width=3.9cm]{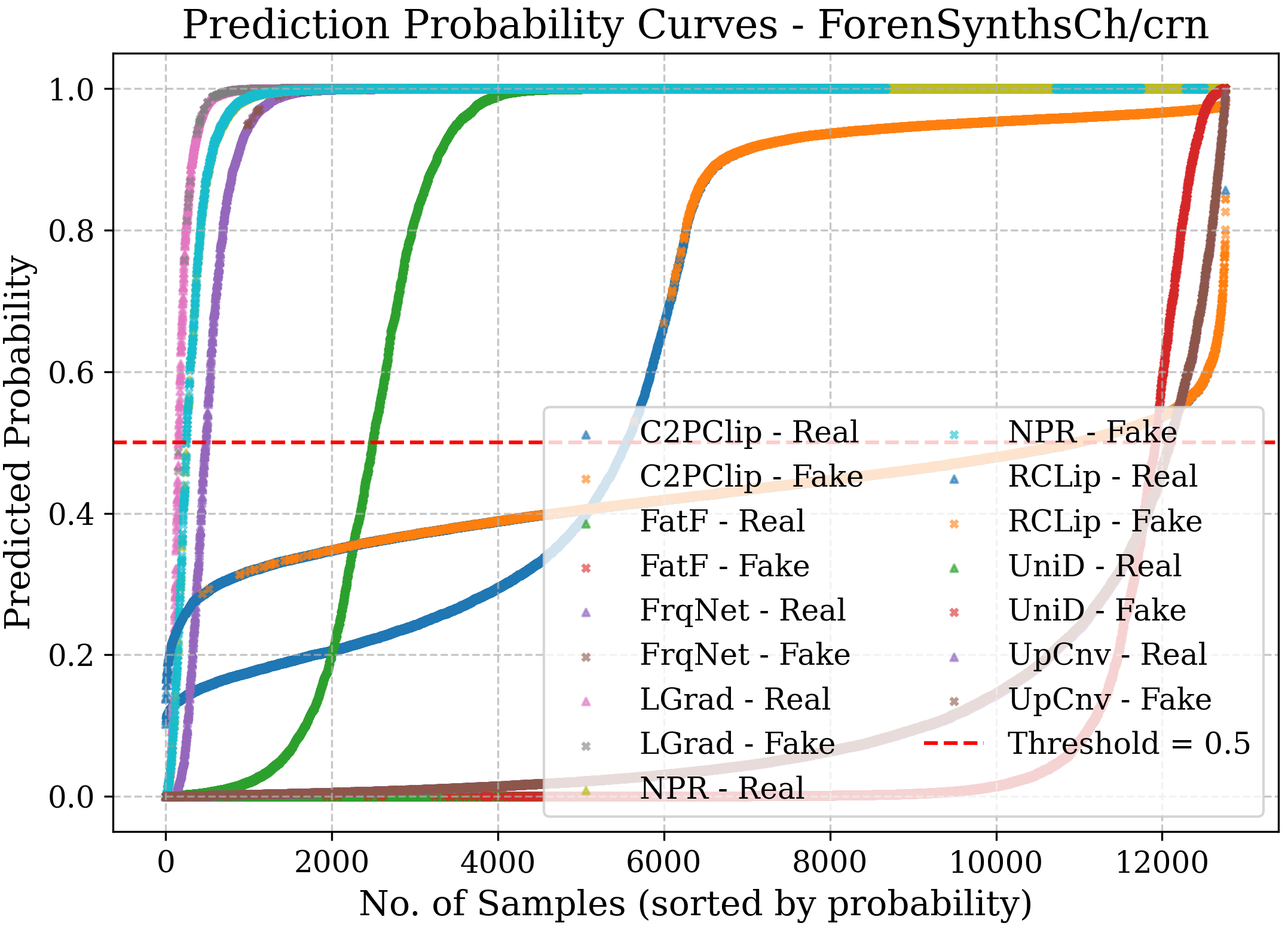} \\
        (a) & (b) & (c) \\  
        \includegraphics[width=3.9cm]{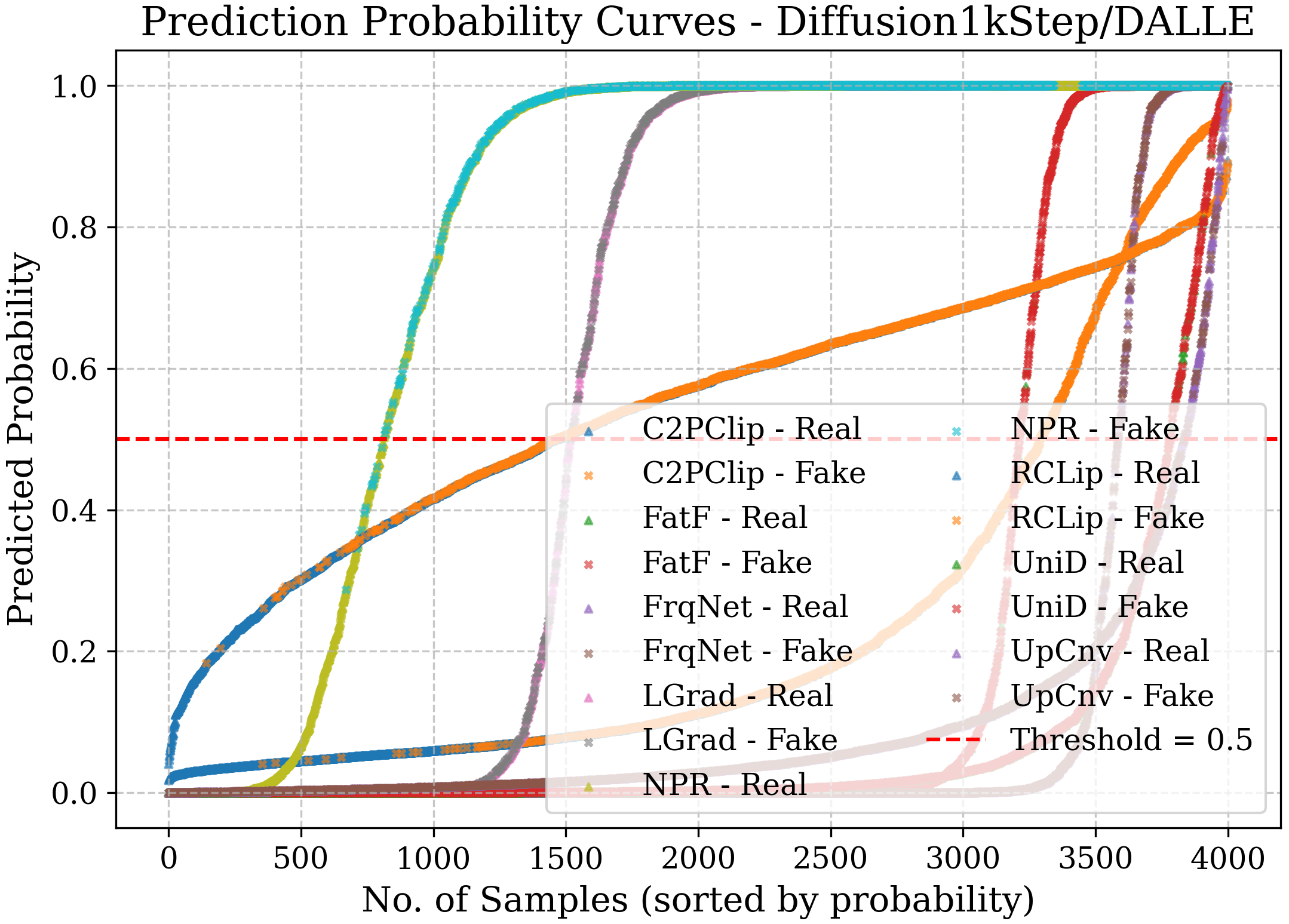} &
        \includegraphics[width=3.9cm]{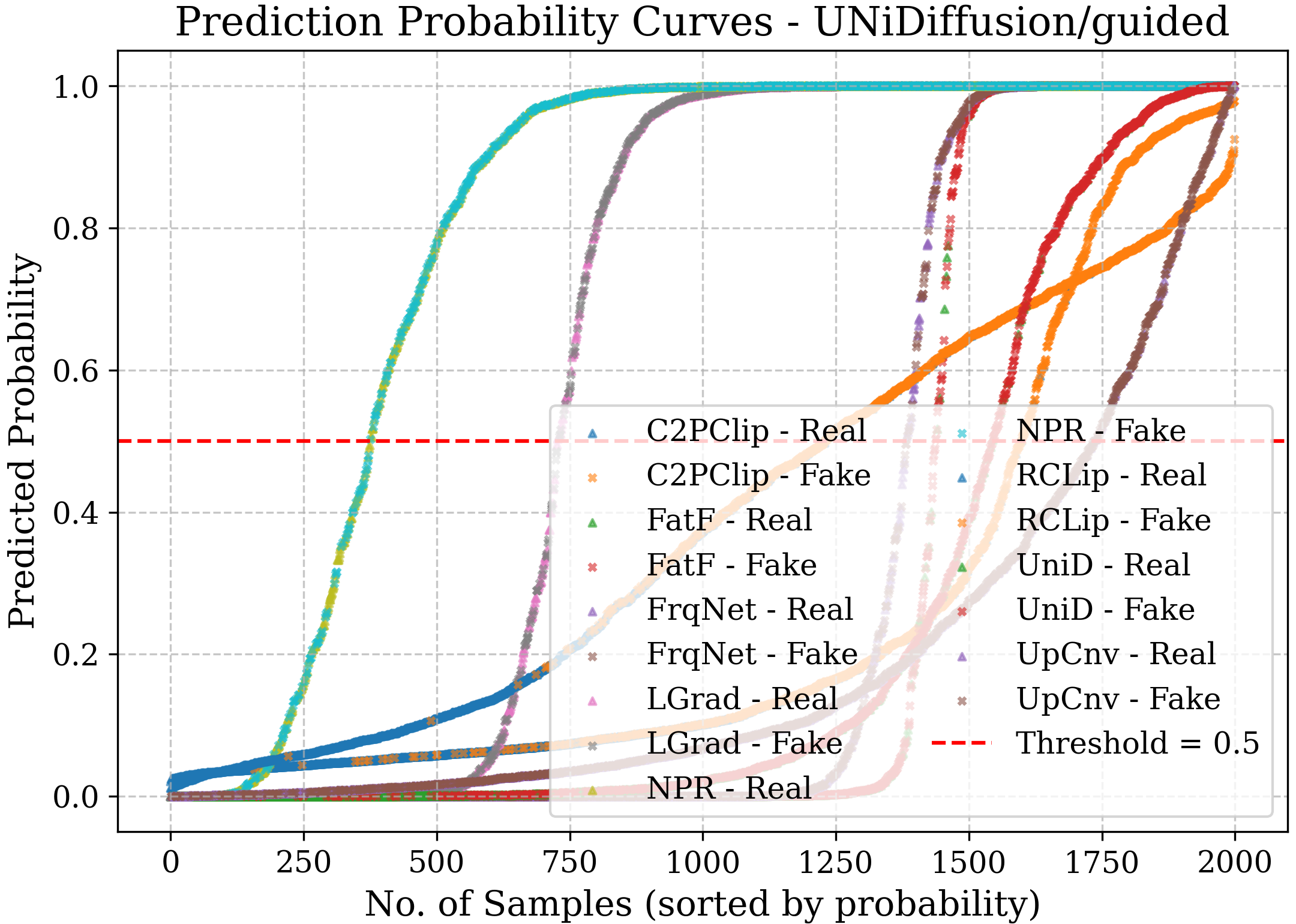} &
        \includegraphics[width=3.9cm]{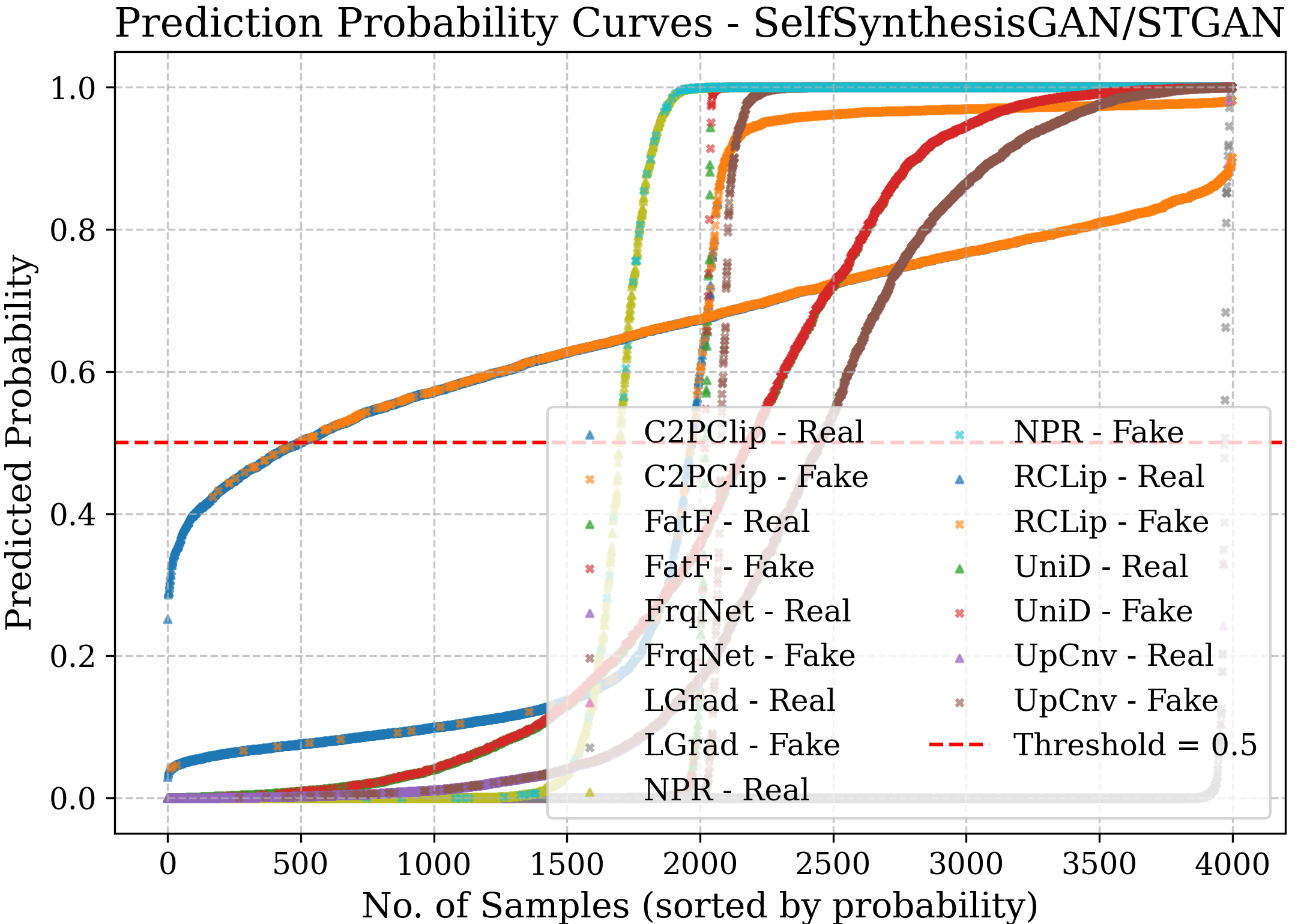} \\
        (d) & (e) & (f)
    \end{tabular}
    \vspace{2pt}
    \caption{Confidence of fake prediction by each model on six datasets: (a) Adm, (b) BigGAN, (c) CRN, (d) Dalle, (e) Guided, and (f) StGAN.}
    \label{fig:conf_curve}
    \vspace{-25pt}
\end{figure}

\vspace{-8pt}
\textbf{Inconsistent experimental settings:} While most methods employ the same training set, variations in their basic experimental configurations lead to inconsistencies across SoTA methods, thereby hindering the reproducibility of results reported in the paper.\\
\textbf{Lack of generalization:} Although most methods claim to generalize to unseen generative models, they struggle with unseen samples, as shown in Tables~\ref{tab:diffusion1kstep}–\ref{tab:mnw}, particularly for the MNW~\cite{MNW2025} dataset in Table~\ref{tab:mnw} while varying the generative models. \\
\textbf{Biases in decision-making:} In many cases, the methods exhibit bias toward either the real or deepfake class. As shown in Table~\ref{tab:mnw}, most methods fail to detect MNW~\cite{MNW2025} samples, whereas NPR~\cite{tan2024rethinking} achieves 91\% ACC. Our analysis of the confidence curve reveals that NPR~\cite{tan2024rethinking} is biased toward fakes, as shown in Figure~\ref{fig:conf_curve}, which enables it to detect the MNW~\cite{MNW2025} dataset. In contrast, UpConv~\cite{durall2020watch} is biased toward real class (Figure~\ref{fig:conf_curve}).\\
\textbf{Restricted preprocessing:} Most methods rely on predefined preprocessing pipelines tailored to specific datasets; for instance, NPR~\cite{tan2024rethinking}, RINE~\cite{koutlis2024leveraging}, and FreqNet~\cite{tan2024frequency} omit cropping for certain datasets, while applying it to others. \\
\textbf{Vulnerability to AFs:} A few studies~\cite{wang2020cnn} have evaluated robustness against conventional AFs, such as noise and compression. However, none have considered AFs based on GANs~\cite{uddin2019anti}, diffusion models~\cite{wang2023dire}, or optimization-based anti-forensic (AF) attacks.\\
\textbf{Lack of explainability:} Most methods lack explainability of their prediction to provide insight into model behavior to make it difficult to understand why a particular decision was made and limiting trust, accountability, and the ability to improve the model effectively.

\subsection{Future Research Directions}\vspace{-8pt}
\textbf{Standardized framework:} Our findings suggest developing unified training, preprocessing, and evaluation protocols to ensure fair comparisons and reproducibility of the results.\\
\textbf{Improved generalization:} To improve generalization to GANs and diffusion, our study suggests domain-agnostic features using meta-learning or multi-domain training.\\
\textbf{Bias reduction:} To better generalize across real and fake classes, our framework recommends balanced objectives and debiasing techniques to avoid skew toward one class.\\
\textbf{Preprocessing robustness:} Additionally, the experimental results encourage building models that work reliably across varied or minimal preprocessing and AF conditions.\\
\textbf{Explainability and trust:} Future research should focus on enhancing model interpretability through explainable AI techniques, such as attention visualization, causal reasoning, rule-based representations, and large language model–driven report generation, to improve trust and usability in real-world forensic applications.\\
\vspace{-22pt}

\vspace{-5pt}
\section{Conclusions}\vspace{-11pt}
In this study, we evaluated SoTA forensic methods under unified configurations across multiple benchmark datasets to reveal their strengths and limitations. Our empirical analysis highlighted key challenges, including inconsistent experimental settings, limited generalization to unseen generative models, biases in decision-making, and dependence on dataset-specific preprocessing. By systematically benchmarking ten SoTA methods across seven datasets, we provided insights into their generalization and applicability in real-world scenarios.\\
Furthermore, we proposed future research directions, including the development of standardized frameworks, improved generalization through domain-agnostic feature learning, bias mitigation strategies, and preprocessing-robust model design. Overall, this study serves as a comprehensive guide for the research community to inspire the development of more robust, generalizable, and explainable approaches for detecting AI-generated media. \textbf{[The code, model weights, and datasets will be released upon acceptance of the paper.]}
\vspace{-15pt}
\bibliography{egbib}

\begin{thebibliography}{64}
\providecommand{\natexlab}[1]{#1}
\providecommand{\url}[1]{\texttt{#1}}
\expandafter\ifx\csname urlstyle\endcsname\relax
  \providecommand{\doi}[1]{doi: #1}\else
  \providecommand{\doi}{doi: \begingroup \urlstyle{rm}\Url}\fi

\bibitem[Agarwal and Farid(2020)]{DeepfakesSocialMedia}
Shruti Agarwal and Hany Farid.
\newblock Detecting deepfake videos from phoneme-viseme mismatches.
\newblock In \emph{Proceedings of the IEEE/CVF Conference on Computer Vision and Pattern Recognition Workshops}, 2020.

\bibitem[Bellemare et~al.()Bellemare, Danihelka, Dabney, Mohamed, Lakshminarayanan, Hoyer, and Munos]{bellemarecramer}
Marc~G Bellemare, Ivo Danihelka, Will Dabney, Shakir Mohamed, Balaji Lakshminarayanan, Stephan Hoyer, and Remi Munos.
\newblock The cramer distance as a solution to biased wasserstein gradients. 2017.
\newblock In \emph{URL https://openreview. net/forum}.

\bibitem[Berthelot et~al.(2017)Berthelot, Schumm, and Metz]{berthelot2017began}
David Berthelot, Thomas Schumm, and Luke Metz.
\newblock Began: Boundary equilibrium generative adversarial networks. arxiv 2017.
\newblock \emph{arXiv preprint arXiv:1703.10717}, 2017.

\bibitem[Brock et~al.(2018)Brock, Donahue, and Simonyan]{brock2018large}
Andrew Brock, Jeff Donahue, and Karen Simonyan.
\newblock Large scale gan training for high fidelity natural image synthesis.
\newblock \emph{arXiv preprint arXiv:1809.11096}, 2018.

\bibitem[{CFO.com}(2024)]{cfo2024}
{CFO.com}.
\newblock Most companies have experienced financial loss due to a deepfake, 2024.
\newblock URL \url{https://www.cfo.com/news/most-companies-have-experienced-financial-loss-due-to-a-deepfake-regula-report/732094/}.
\newblock [Online; accessed 2024].

\bibitem[Chen and Koltun(2017)]{chen2017photographic}
Qifeng Chen and Vladlen Koltun.
\newblock Photographic image synthesis with cascaded refinement networks.
\newblock In \emph{Proceedings of the IEEE international conference on computer vision}, pages 1511--1520, 2017.

\bibitem[Chesney and Citron(2019)]{DeepfakeNews}
Robert Chesney and Danielle Citron.
\newblock Deep fakes: A looming challenge for privacy, democracy, and national security.
\newblock \emph{California Law Review}, 107\penalty0 (1):\penalty0 175--200, 2019.

\bibitem[Choi et~al.(2018)Choi, Choi, Kim, Ha, Kim, and Choo]{choi2018stargan}
Yunjey Choi, Minje Choi, Munyoung Kim, Jung-Woo Ha, Sunghun Kim, and Jaegul Choo.
\newblock Stargan: Unified generative adversarial networks for multi-domain image-to-image translation.
\newblock In \emph{Proceedings of the IEEE conference on computer vision and pattern recognition}, pages 8789--8797, 2018.

\bibitem[Choo(2019)]{IoT}
Kim-Kwang~Raymond Choo.
\newblock The dangers of fake content in the age of iot and ai.
\newblock \emph{Computer Fraud \& Security}, 2019\penalty0 (5):\penalty0 10--13, 2019.

\bibitem[Cozzolino et~al.(2024)Cozzolino, Poggi, Corvi, Nie{\ss}ner, and Verdoliva]{cozzolino2024raising}
Davide Cozzolino, Giovanni Poggi, Riccardo Corvi, Matthias Nie{\ss}ner, and Luisa Verdoliva.
\newblock Raising the bar of ai-generated image detection with clip (2023).
\newblock \emph{arXiv preprint arXiv:2312.00195}, 2024.

\bibitem[Dai et~al.(2019)Dai, Cai, Zhang, Xia, and Zhang]{dai2019second}
Tao Dai, Jianrui Cai, Yongbing Zhang, Shu-Tao Xia, and Lei Zhang.
\newblock Second-order attention network for single image super-resolution.
\newblock In \emph{Proceedings of the IEEE/CVF conference on computer vision and pattern recognition}, pages 11065--11074, 2019.

\bibitem[Dhariwal and Nichol(2021)]{dhariwal2021diffusion}
Prafulla Dhariwal and Alexander Nichol.
\newblock Diffusion models beat gans on image synthesis.
\newblock \emph{Advances in neural information processing systems}, 34:\penalty0 8780--8794, 2021.

\bibitem[Durall et~al.(2020)Durall, Keuper, and Keuper]{durall2020watch}
Ricard Durall, Margret Keuper, and Janis Keuper.
\newblock Watch your up-convolution: Cnn based generative deep neural networks are failing to reproduce spectral distributions.
\newblock In \emph{Proceedings of the IEEE/CVF conference on computer vision and pattern recognition}, pages 7890--7899, 2020.

\bibitem[{Eftsure}(2024)]{eftsure2024}
{Eftsure}.
\newblock Statistics: Deepfake fraud in 2024, 2024.
\newblock URL \url{https://www.eftsure.com/statistics/deepfake-statistics/}.
\newblock [Online; accessed 2024].

\bibitem[{Eftsure}(2025)]{eftsureForecast}
{Eftsure}.
\newblock Deepfake-related fraud forecast to hit \$40b by 2027, 2025.
\newblock URL \url{https://www.eftsure.com/statistics/deepfake-statistics/?utm_source=chatgpt.com}.
\newblock [Online; accessed 2025].

\bibitem[{eSecurityPlanet.com}(2025)]{esecurityplanet2025}
{eSecurityPlanet.com}.
\newblock {AI Deepfakes Surge: \$200 Million Lost}, 2025.
\newblock URL \url{https://www.esecurityplanet.com/news/ai-deepfakes-surge-200-million-lost/}.
\newblock [Online; accessed 2025].

\bibitem[Frank et~al.(2020)Frank, Eisenhofer, Sch{\"o}nherr, Fischer, Kolossa, and Holz]{frank2020leveraging}
Joshua Frank, Thorsten Eisenhofer, Lea Sch{\"o}nherr, Andreas Fischer, Dorothea Kolossa, and Thorsten Holz.
\newblock Leveraging frequency analysis for deep fake image recognition.
\newblock In \emph{International Conference on Machine Learning}, pages 3247--3258. PMLR, 2020.

\bibitem[Fridman et~al.(2020)Fridman, Brown, and Mericli]{AutonomousDriving}
Jason Fridman, John Brown, and Can Mericli.
\newblock Synthetic video attacks on autonomous driving systems using gans.
\newblock \emph{arXiv preprint arXiv:2001.03667}, 2020.

\bibitem[{Globe Newswire}(2024)]{globenewswire2024}
{Globe Newswire}.
\newblock Deepfake fraud costs the financial sector an average of \$600,000 per company, 2024.
\newblock URL \url{https://www.businesswire.com/news/home/20241031656724/en/Deepfake-Fraud-Costs}.
\newblock [Online; accessed 2024].

\bibitem[Gu et~al.(2022)Gu, Chen, Bao, Wen, Zhang, Chen, Yuan, and Guo]{gu2022vector}
Shuyang Gu, Dong Chen, Jianmin Bao, Fang Wen, Bo~Zhang, Dongdong Chen, Lu~Yuan, and Baining Guo.
\newblock Vector quantized diffusion model for text-to-image synthesis.
\newblock In \emph{Proceedings of the IEEE/CVF conference on computer vision and pattern recognition}, pages 10696--10706, 2022.

\bibitem[He et~al.(2019)He, Zuo, Kan, Shan, and Chen]{he2019attgan}
Zhenliang He, Wangmeng Zuo, Meina Kan, Shiguang Shan, and Xilin Chen.
\newblock Attgan: Facial attribute editing by only changing what you want.
\newblock \emph{IEEE transactions on image processing}, 28\penalty0 (11):\penalty0 5464--5478, 2019.

\bibitem[Ho et~al.(2020)Ho, Jain, and Abbeel]{ho2020denoising}
Jonathan Ho, Ajay Jain, and Pieter Abbeel.
\newblock Denoising diffusion probabilistic models.
\newblock \emph{Advances in neural information processing systems}, 33:\penalty0 6840--6851, 2020.

\bibitem[Karras et~al.(2017)Karras, Aila, Laine, and Lehtinen]{karras2017progressive}
Tero Karras, Timo Aila, Samuli Laine, and Jaakko Lehtinen.
\newblock Progressive growing of gans for improved quality, stability, and variation.
\newblock \emph{arXiv preprint arXiv:1710.10196}, 2017.

\bibitem[Karras et~al.(2019)Karras, Laine, and Aila]{karras2019style}
Tero Karras, Samuli Laine, and Timo Aila.
\newblock A style-based generator architecture for generative adversarial networks.
\newblock In \emph{Proceedings of the IEEE/CVF conference on computer vision and pattern recognition}, pages 4401--4410, 2019.

\bibitem[Karras et~al.(2020)Karras, Laine, Aittala, Hellsten, Lehtinen, and Aila]{karras2020analyzing}
Tero Karras, Samuli Laine, Miika Aittala, Janne Hellsten, Jaakko Lehtinen, and Timo Aila.
\newblock Analyzing and improving the image quality of stylegan.
\newblock In \emph{Proceedings of the IEEE/CVF conference on computer vision and pattern recognition}, pages 8110--8119, 2020.

\bibitem[Korshunov and Marcel(2018)]{Surveillance}
Pavel Korshunov and S{\'e}bastien Marcel.
\newblock Deepfakes: a new threat to face recognition? assessment and detection.
\newblock \emph{arXiv preprint arXiv:1812.08685}, 2018.

\bibitem[Koutlis and Papadopoulos(2024)]{koutlis2024leveraging}
Christos Koutlis and Symeon Papadopoulos.
\newblock Leveraging representations from intermediate encoder-blocks for synthetic image detection.
\newblock In \emph{European Conference on Computer Vision}, pages 394--411. Springer, 2024.

\bibitem[Lee et~al.(2021)Lee, Tran, and Cheung]{lee2021infomax}
Kwot~Sin Lee, Ngoc-Trung Tran, and Ngai-Man Cheung.
\newblock Infomax-gan: Improved adversarial image generation via information maximization and contrastive learning.
\newblock In \emph{Proceedings of the IEEE/CVF winter conference on applications of computer vision}, pages 3942--3952, 2021.

\bibitem[Li et~al.(2017)Li, Chang, Cheng, Yang, and P{\'o}czos]{li2017mmd}
Chun-Liang Li, Wei-Cheng Chang, Yu~Cheng, Yiming Yang, and Barnab{\'a}s P{\'o}czos.
\newblock Mmd gan: Towards deeper understanding of moment matching network.
\newblock \emph{Advances in neural information processing systems}, 30, 2017.

\bibitem[Li et~al.(2019)Li, Zhang, and Malik]{li2019diverse}
Ke~Li, Tianhao Zhang, and Jitendra Malik.
\newblock Diverse image synthesis from semantic layouts via conditional imle. 2019 ieee.
\newblock In \emph{CVF International Conference on Computer Vision (ICCV)}, pages 4219--4228, 2019.

\bibitem[Liu et~al.(2024)Liu, Tan, Tan, Wei, Wang, and Zhao]{liu2024forgery}
Huan Liu, Zichang Tan, Chuangchuang Tan, Yunchao Wei, Jingdong Wang, and Yao Zhao.
\newblock Forgery-aware adaptive transformer for generalizable synthetic image detection.
\newblock In \emph{Proceedings of the IEEE/CVF Conference on Computer Vision and Pattern Recognition}, pages 10770--10780, 2024.

\bibitem[Liu et~al.(2022)Liu, Ren, Lin, and Zhao]{liu2022pseudo}
Luping Liu, Yi~Ren, Zhijie Lin, and Zhou Zhao.
\newblock Pseudo numerical methods for diffusion models on manifolds.
\newblock \emph{arXiv preprint arXiv:2202.09778}, 2022.

\bibitem[Liu et~al.(2019)Liu, Ding, Xia, Liu, Ding, Zuo, and Wen]{liu2019stgan}
Ming Liu, Yukang Ding, Min Xia, Xiao Liu, Errui Ding, Wangmeng Zuo, and Shilei Wen.
\newblock Stgan: A unified selective transfer network for arbitrary image attribute editing.
\newblock In \emph{Proceedings of the IEEE/CVF conference on computer vision and pattern recognition}, pages 3673--3682, 2019.

\bibitem[Lu{\v{c}}i{\'c} et~al.(2019)Lu{\v{c}}i{\'c}, Tschannen, Ritter, Zhai, Bachem, and Gelly]{luvcic2019high}
Mario Lu{\v{c}}i{\'c}, Michael Tschannen, Marvin Ritter, Xiaohua Zhai, Olivier Bachem, and Sylvain Gelly.
\newblock High-fidelity image generation with fewer labels.
\newblock In \emph{International conference on machine learning}, pages 4183--4192. PMLR, 2019.

\bibitem[Miyato et~al.(2018)Miyato, Kataoka, Koyama, and Yoshida]{miyato2018spectral}
Takeru Miyato, Toshiki Kataoka, Masanori Koyama, and Yuichi Yoshida.
\newblock Spectral normalization for generative adversarial networks.
\newblock \emph{arXiv preprint arXiv:1802.05957}, 2018.

\bibitem[Nichol et~al.(2021)Nichol, Dhariwal, Ramesh, Shyam, Mishkin, McGrew, Sutskever, and Chen]{nichol2021glide}
Alex Nichol, Prafulla Dhariwal, Aditya Ramesh, Pranav Shyam, Pamela Mishkin, Bob McGrew, Ilya Sutskever, and Mark Chen.
\newblock Glide: Towards photorealistic image generation and editing with text-guided diffusion models.
\newblock \emph{arXiv preprint arXiv:2112.10741}, 2021.

\bibitem[Nichol and Dhariwal(2021)]{nichol2021improved}
Alexander~Quinn Nichol and Prafulla Dhariwal.
\newblock Improved denoising diffusion probabilistic models.
\newblock In \emph{International conference on machine learning}, pages 8162--8171. PMLR, 2021.

\bibitem[Nie et~al.(2018)Nie, Narodytska, and Patel]{nie2018relgan}
Weili Nie, Nina Narodytska, and Ankit Patel.
\newblock Relgan: Relational generative adversarial networks for text generation.
\newblock In \emph{International conference on learning representations}, 2018.

\bibitem[Ojha et~al.(2023)Ojha, Li, and Lee]{ojha2023towards}
Utkarsh Ojha, Yuheng Li, and Yong~Jae Lee.
\newblock Towards universal fake image detectors that generalize across generative models.
\newblock In \emph{Proceedings of the IEEE/CVF Conference on Computer Vision and Pattern Recognition}, pages 24480--24489, 2023.

\bibitem[Park et~al.(2019)Park, Liu, Wang, and Zhu]{park2019semantic}
Taesung Park, Ming-Yu Liu, Ting-Chun Wang, and Jun-Yan Zhu.
\newblock Semantic image synthesis with spatially-adaptive normalization.
\newblock In \emph{Proceedings of the IEEE/CVF conference on computer vision and pattern recognition}, pages 2337--2346, 2019.

\bibitem[Ramesh et~al.(2021)Ramesh, Pavlov, Goh, Gray, Voss, Radford, Chen, and Sutskever]{ramesh2021zero}
Aditya Ramesh, Mikhail Pavlov, Gabriel Goh, Scott Gray, Chelsea Voss, Alec Radford, Mark Chen, and Ilya Sutskever.
\newblock Zero-shot text-to-image generation.
\newblock In \emph{International conference on machine learning}, pages 8821--8831. Pmlr, 2021.

\bibitem[{Reality Defender}(2023)]{realitydefender2023}
{Reality Defender}.
\newblock Understanding the hidden costs of deepfake fraud in finance, 2023.
\newblock URL \url{https://www.realitydefender.com/insights/understanding-the-hidden-costs-of-deepfake-fraud-in-finance}.
\newblock [Online; accessed 2023].

\bibitem[{Reddit r/SocialEngineering}(2025)]{reddit2025}
{Reddit r/SocialEngineering}.
\newblock Deepfake-related crypto scams statistics, 2025.
\newblock URL \url{https://www.reddit.com/r/SocialEngineering/comments/1hwxvhp/how_are_scammers_using_5_deepfakes_to_steal/}.
\newblock [Online; accessed 2025].

\bibitem[Roca et~al.(2025)Roca, Postiglione, Gao, Gortner, Wojciak, Wang, Alimardani, Anlen, White, Lavista, Kraus, Gregory, and Subrahmanian]{MNW2025}
Thomas Roca, Marco Postiglione, Chongyang Gao, Isabel Gortner, Zuzanna Wojciak, Pengce Wang, Masah Alimardani, Shirin Anlen, Kevin White, Juan Lavista, Sarit Kraus, Sam Gregory, and V.S. Subrahmanian.
\newblock Introducing the mnw benchmark for ai forensics.
\newblock \url{https://github.com/nsail-lab/MNW}, 2025.
\newblock PDF.

\bibitem[Rombach et~al.(2022)Rombach, Blattmann, Lorenz, Esser, and Ommer]{rombach2022high}
Robin Rombach, Andreas Blattmann, Dominik Lorenz, Patrick Esser, and Bj{\"o}rn Ommer.
\newblock High-resolution image synthesis with latent diffusion models.
\newblock In \emph{Proceedings of the IEEE/CVF conference on computer vision and pattern recognition}, pages 10684--10695, 2022.

\bibitem[Rossler et~al.(2019)Rossler, Cozzolino, Verdoliva, Riess, Thies, and Nie{\ss}ner]{rossler2019faceforensics}
Andreas Rossler, Davide Cozzolino, Luisa Verdoliva, Christian Riess, Justus Thies, and Matthias Nie{\ss}ner.
\newblock Faceforensics++: Learning to detect manipulated facial images.
\newblock In \emph{Proceedings of the IEEE/CVF international conference on computer vision}, pages 1--11, 2019.

\bibitem[Russakovsky et~al.(2015)Russakovsky, Deng, Su, Krause, Satheesh, Ma, Huang, Karpathy, Khosla, Bernstein, et~al.]{russakovsky2015imagenet}
Olga Russakovsky, Jia Deng, Hao Su, Jonathan Krause, Sanjeev Satheesh, Sean Ma, Zhiheng Huang, Andrej Karpathy, Aditya Khosla, Michael Bernstein, et~al.
\newblock Imagenet large scale visual recognition challenge.
\newblock \emph{International journal of computer vision}, 115\penalty0 (3):\penalty0 211--252, 2015.

\bibitem[Tan et~al.(2025)Tan, Chuangchuang, Renshuai, Huan, Guanghua, Baoyuan, and Yunchao]{c2pclip2025}
Tan, Tao Chuangchuang, Liu Renshuai, Gu~Huan, Wu~Guanghua, Wei Baoyuan, Zhao nad~Yao, and Yunchao.
\newblock C2p-clip: Injecting category common prompt in clip to enhance generalization in deepfake detection.
\newblock In \emph{Proceedings of the AAAI Conference on Artificial Intelligence}, pages 7184--7192, 2025.

\bibitem[Tan et~al.(2023)Tan, Zhao, Wei, Gu, and Wei]{tan2023learning}
Chuangchuang Tan, Yao Zhao, Shikui Wei, Guanghua Gu, and Yunchao Wei.
\newblock Learning on gradients: Generalized artifacts representation for gan-generated images detection.
\newblock In \emph{Proceedings of the IEEE/CVF Conference on Computer Vision and Pattern Recognition}, pages 12105--12114, 2023.

\bibitem[Tan et~al.(2024{\natexlab{a}})Tan, Zhao, Wei, Gu, Liu, and Wei]{tan2024frequency}
Chuangchuang Tan, Yao Zhao, Shikui Wei, Guanghua Gu, Ping Liu, and Yunchao Wei.
\newblock Frequency-aware deepfake detection: Improving generalizability through frequency space domain learning.
\newblock In \emph{Proceedings of the AAAI Conference on Artificial Intelligence}, pages 5052--5060, 2024{\natexlab{a}}.

\bibitem[Tan et~al.(2024{\natexlab{b}})Tan, Zhao, Wei, Gu, Liu, and Wei]{tan2024rethinking}
Chuangchuang Tan, Yao Zhao, Shikui Wei, Guanghua Gu, Ping Liu, and Yunchao Wei.
\newblock Rethinking the up-sampling operations in cnn-based generative network for generalizable deepfake detection.
\newblock In \emph{Proceedings of the IEEE/CVF Conference on Computer Vision and Pattern Recognition}, pages 28130--28139, 2024{\natexlab{b}}.

\bibitem[Tasnim and Baek(2022)]{tasnim2022deep}
Nusrat Tasnim and Joong-Hwan Baek.
\newblock Deep learning-based human action recognition with key-frames sampling using ranking methods.
\newblock \emph{Applied Sciences}, 12\penalty0 (9):\penalty0 4165, 2022.

\bibitem[Tasnim and Baek(2023)]{tasnim2023dynamic}
Nusrat Tasnim and Joong-Hwan Baek.
\newblock Dynamic edge convolutional neural network for skeleton-based human action recognition.
\newblock \emph{Sensors}, 23\penalty0 (2):\penalty0 778, 2023.

\bibitem[Tasnim et~al.(2020)Tasnim, Islam, and Baek]{tasnim2020deep}
Nusrat Tasnim, Md~Mahbubul Islam, and Joong-Hwan Baek.
\newblock Deep learning-based action recognition using 3d skeleton joints information.
\newblock \emph{Inventions}, 5\penalty0 (3):\penalty0 49, 2020.

\bibitem[Uddin et~al.(2019)Uddin, Yang, and Oh]{uddin2019anti}
Kutub Uddin, Yoonmo Yang, and Byung~Tae Oh.
\newblock Anti-forensic against double jpeg compression detection using adversarial generative network.
\newblock \emph{In Proceedings of the Korean Society of Broadcast Engineers Conference}, pages 58--60, 2019.

\bibitem[Uddin et~al.(2023)Uddin, Yang, Jeong, and Oh]{uddin2023robust}
Kutub Uddin, Yoonmo Yang, Tae~Hyun Jeong, and Byung~Tae Oh.
\newblock A robust open-set multi-instance learning for defending adversarial attacks in digital image.
\newblock \emph{IEEE Transactions on Information Forensics and Security}, 19:\penalty0 2098--2111, 2023.

\bibitem[Uddin et~al.(2024)Uddin, Jeong, and Oh]{uddin2024counter}
Kutub Uddin, Tae~Hyun Jeong, and Byung~Tae Oh.
\newblock Counter-act against gan-based attacks: A collaborative learning approach for anti-forensic detection.
\newblock \emph{Applied Soft Computing}, 153:\penalty0 111287, 2024.

\bibitem[Uddin et~al.(2025{\natexlab{a}})Uddin, Khan, Farooq, and Malik]{uddin2025shield}
Kutub Uddin, Awais Khan, Muhammad~Umar Farooq, and Khalid Malik.
\newblock Shield: A secure and highly enhanced integrated learning for robust deepfake detection against adversarial attacks.
\newblock \emph{arXiv preprint arXiv:2507.13170}, 2025{\natexlab{a}}.

\bibitem[Uddin et~al.(2025{\natexlab{b}})Uddin, Tasnim, Saeed, and Malik]{uddin2025guard}
Kutub Uddin, Nusrat Tasnim, Muhammad~Saad Saeed, and Khalid~Mahmood Malik.
\newblock Guard: Generative unmasking and adversarial-resistant deepfake detection using multi-model knowledge distillation.
\newblock \emph{Authorea Preprints}, 2025{\natexlab{b}}.

\bibitem[Wang et~al.(2020)Wang, Wang, Zhang, Owens, and Efros]{wang2020cnn}
Sheng-Yu Wang, Oliver Wang, Richard Zhang, Andrew Owens, and Alexei~A Efros.
\newblock Cnn-generated images are surprisingly easy to spot... for now.
\newblock In \emph{Proceedings of the IEEE/CVF conference on computer vision and pattern recognition}, pages 8695--8704, 2020.

\bibitem[Wang et~al.(2023)Wang, Bao, Zhou, Wang, Hu, Chen, and Li]{wang2023dire}
Zhendong Wang, Jianmin Bao, Wengang Zhou, Weilun Wang, Hezhen Hu, Hong Chen, and Houqiang Li.
\newblock Dire for diffusion-generated image detection.
\newblock In \emph{Proceedings of the IEEE/CVF International Conference on Computer Vision}, pages 22445--22455, 2023.

\bibitem[Yang et~al.(2019)Yang, Li, and Lyu]{RemoteAccess}
Xin Yang, Yuezun Li, and Siwei Lyu.
\newblock Exposing deep fakes using inconsistent head poses.
\newblock In \emph{Proceedings of the IEEE International Conference on Acoustics, Speech and Signal Processing (ICASSP)}, 2019.

\bibitem[Yi et~al.(2019)Yi, Liu, Zhang, and Tan]{BiometricThreat}
Zhen Yi, Qiang Liu, Yuan Zhang, and Li~Tan.
\newblock Deep learning based face recognition: A survey.
\newblock \emph{IEEE Access}, 7:\penalty0 106395--106413, 2019.

\bibitem[Zhu et~al.(2017)Zhu, Park, Isola, and Efros]{zhu2017unpaired}
Jun-Yan Zhu, Taesung Park, Phillip Isola, and Alexei~A Efros.
\newblock Unpaired image-to-image translation using cycle-consistent adversarial networks.
\newblock In \emph{Proceedings of the IEEE international conference on computer vision}, pages 2223--2232, 2017.

\end{thebibliography}
\end{document}